\documentclass{article}

 \usepackage[main, preprint]{neurips_2026}

\usepackage[utf8]{inputenc} 
\usepackage[T1]{fontenc}    
\usepackage{hyperref}       
\usepackage{url}            
\usepackage{booktabs}       
\usepackage{amsfonts}       
\usepackage{nicefrac}       
\usepackage{microtype}      
\usepackage{xcolor}         
\usepackage{comment}         

\usepackage{lipsum}
\usepackage{amsmath}
\usepackage{multirow}
\usepackage{graphicx}
\usepackage{makecell}
\newcommand{\ours}{MVTrack4Gen}
\usepackage{wrapfig}  
\usepackage{bbm}
\usepackage{xcolor}
\definecolor{impgreen}{RGB}{0,150,0}
\definecolor{impred}{RGB}{200,0,0}

\newcommand{\gain}[1]{{\tiny\textcolor{impgreen}{(#1)}}}
\newcommand{\loss}[1]{{\tiny\textcolor{impred}{(#1)}}}

\definecolor{refgreen}{RGB}{0,150,0}
\definecolor{genred}{RGB}{200,0,0}

\title{MVTrack4Gen: Multi-View Point Tracking as Geometric Supervision for 4D Video Generation
}

\author{JoungBin Lee\textsuperscript{1} \quad Jaewoo Jung\textsuperscript{1} \quad Jongmin Lee\textsuperscript{1} \quad Tongmin Kim\textsuperscript{1} \quad Hyunsung Kim\textsuperscript{1} \\
\textbf{Takuya Narihira}\textsuperscript{2} \quad \textbf{Kazumi Fukuda}\textsuperscript{2} \quad \textbf{Jahyeok Koo}\textsuperscript{1} \quad \textbf{Jisang Han}\textsuperscript{1} \\
\textbf{Yuki Mitsufuji}\textsuperscript{2,3}$^\dagger$ \quad \textbf{Seungryong Kim}\textsuperscript{1}$^\dagger$ \\[2pt]
\textsuperscript{1}KAIST AI \quad \textsuperscript{2}Sony AI \quad \textsuperscript{3}Sony Group Corporation \\[2pt]
\textbf{Project Page:} \href{https://cvlab-kaist.github.io/MVTrack4Gen/}{\texttt{https://cvlab-kaist.github.io/MVTrack4Gen/}}
\vspace{-10pt} \\
}

\begin{document}
\footnotetext[2]{Corresponding authors.}

\maketitle

\begin{center}
    \begin{figure}[h]
        \centering
        \vspace{-10pt}
        \includegraphics[width=0.95\linewidth]{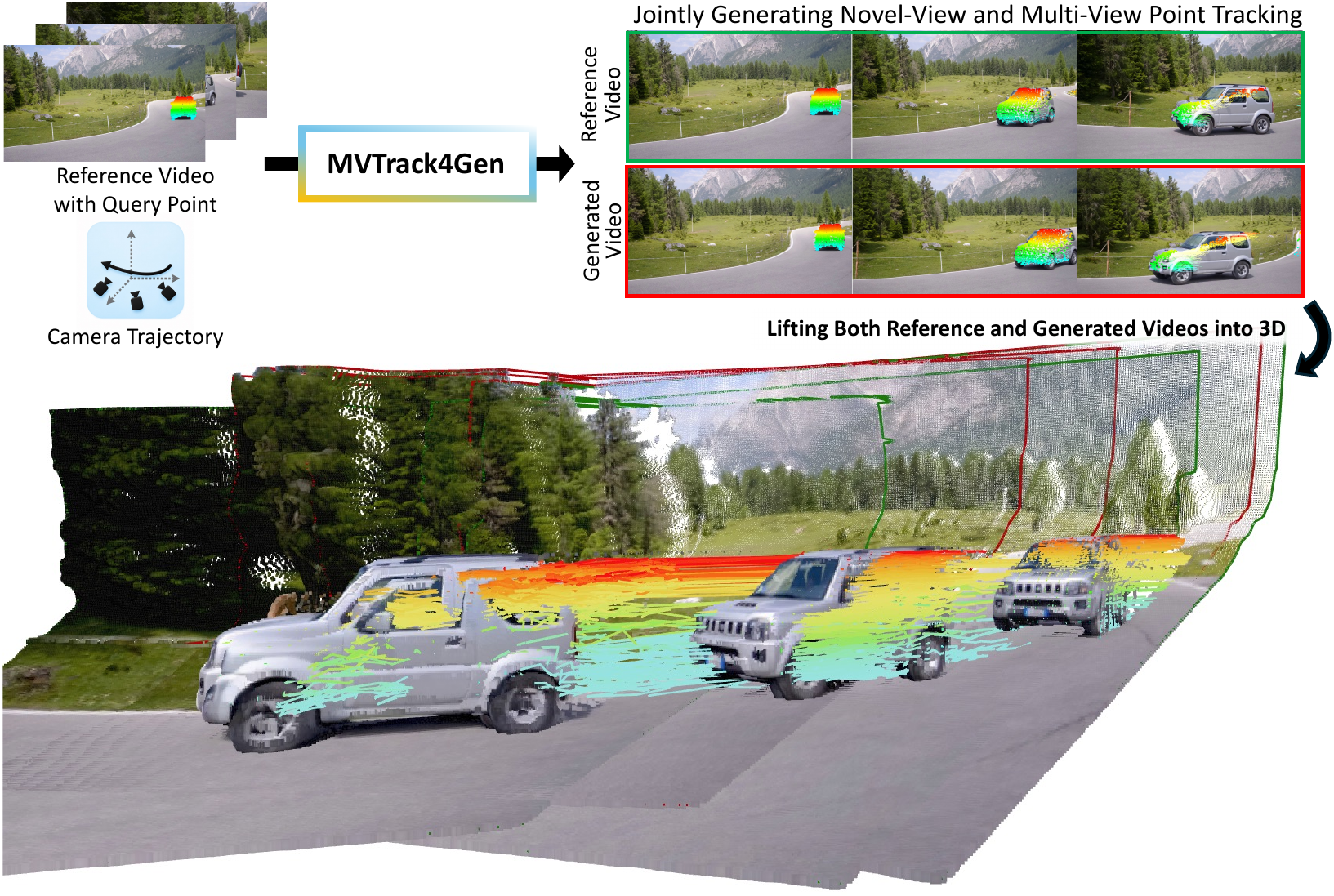}
        \vspace{-10pt}
        \caption{
           \textbf{\ours}\ jointly generates a novel-view video and multi-view point tracks, given a monocular reference video with query points and a user-specified camera trajectory. Lifting both the \textcolor{refgreen}{\textbf{reference}} and \textcolor{genred}{\textbf{generated}} frames into 3D space using Depth Anything 3~\cite{lin2025depth} shows that the target views and multi-view point tracks faithfully preserve the dynamic motion of the reference video while remaining geometrically consistent.
        }
        \vspace{-20pt}
        \label{fig:teaser}
    \end{figure}
\end{center}    

\begin{abstract}

Synthesizing a novel-view video from a monocular reference video along a target camera trajectory requires both geometric consistency and motion fidelity with respect to the reference video. Existing methods based on explicit 3D representations are limited by the accuracy of off-the-shelf reconstruction modules, which often produce inaccurate geometry for dynamic objects in monocular videos. In contrast, camera-conditioning-only methods can achieve high visual quality but often struggle to preserve geometric and motion consistency. In this work, we introduce \textbf{\ours} (\textbf{M}ulti-\textbf{V}iew point \textbf{Track}ing \textbf{for} Novel-View \textbf{Gen}eration), a motion-aware training framework that leverages multi-view point tracking as an additional geometric and motion supervision signal for camera-conditioning-only novel-view video diffusion models. Our key finding is that specific attention layers encode strong correspondence cues, where query features attend to key features at geometrically corresponding locations across views and over time, and the misalignment of these correspondences causes motion inconsistency. Based on this observation, we route these features into an auxiliary multi-view tracking head and jointly train the diffusion model with a point-tracking objective. By explicitly strengthening these motion-aware correspondences, \textbf{\ours} improves existing models to better follow the motion in the reference view and maintain cross-view geometric consistency. Across diverse benchmarks, our method achieves state-of-the-art geometric consistency and competitive camera accuracy.

\end{abstract}

\section{Introduction}
Novel-view video generation from a monocular video aims to synthesize a target-view video along a user-specified camera trajectory, with applications in virtual cinematography, robotics, and immersive AR/VR. To be useful in these real-world scenarios, the generated video should satisfy four requirements: (1) accurate camera control that faithfully follows the user-specified trajectory, (2) geometric consistency in which the scene structure of the reference video is preserved across the synthesized view, (3) motion consistency that maintains the dynamics of the reference video, and (4) a photorealistic visual appearance. The central challenge is achieving all four at once, since the model must respect the underlying 3D geometry and motion of the scene while still producing high-quality videos.

Recent works leverage generative priors from pretrained video diffusion models~\cite{agarwal2025cosmos, bar2024lumiere, kong2024hunyuanvideo, wan2025wan, yang2025cogvideox} to address this task, achieving notable progress in dynamic scenes~\cite{ren2025gen3c, lee20253d, yu2025trajectorycrafter, chen2025postcam, yang2026neoverse, bai2025recammaster, park2025redirector, huang2025spacetimepilot}. These approaches largely fall into two paradigms. The first paradigm follows a \textit{reconstruct-then-generate} pipeline~\cite{ren2025gen3c, lee20253d, yu2025trajectorycrafter, chen2025postcam, yang2026neoverse}, which first reconstructs an explicit 3D representation from the reference video, projects this geometry along the target camera trajectory, and feeds the projected appearance as a spatial condition. By supplying explicit geometric guidance, this design relieves the model from having to jointly infer geometry and synthesize photorealistic frames. However, its quality hinges on the accuracy of the reconstructed geometry, which becomes challenging for dynamic scenes, where inaccurate dynamic-object geometry introduces distortions and flying-pixel artifacts near object boundaries. These artifacts corrupt the spatial condition and prevent the diffusion model from faithfully capturing the geometry and motion of dynamic objects.

To avoid this dependence on error-prone reconstruction, a second and more recent line of work performs \textit{camera-conditioning-only} generation~\cite{bai2025recammaster, park2025redirector, huang2025spacetimepilot}, without any explicit 3D representation. These methods feed the reference video to the diffusion model as additional input tokens and inject the target trajectory through camera embeddings such as Plücker coordinates~\cite{sitzmann2021light}, and can be trained from pairs of multi-view videos. Because the reference and target views are processed jointly within the same 3D attention modules, cross-view and intra-view information is exchanged implicitly at every layer. This implicit design yields markedly more photorealistic results, but the lack of explicit geometric grounding leaves the model with a weak understanding of scene structure. As a result, dynamic objects are often placed at incorrect locations or assigned inconsistent motions, producing cross-view geometric and motion inconsistencies.

In this work, we start from the observation that camera-conditioning-only methods already excel at photorealistic synthesis, and ask whether their geometric understanding can be improved without reintroducing explicit 3D conditioning. To this end, we analyze the 3D attention maps of these diffusion models~\cite{bai2025recammaster, park2025redirector} across layers and denoising timesteps, where information is implicitly exchanged across views and frames. Inspired by~\cite{nam2025emergenttemporalcorrespondencesvideo}, which reveals emergent correspondence in the attention maps of video diffusion models, our analysis yields three key observations. First, query-key matching within the 3D attention blocks provides clear correspondence cues, capturing both intra-video temporal correspondences within each view and inter-video cross-view correspondences between the generated and reference views. Second, these temporal and cross-view correspondences become simultaneously prominent at specific intermediate layers, revealing which regions the model attends to when synthesizing each part of the generated frame. Third, in regions where dynamic objects exhibit geometric or motion inconsistencies, the attention maps at these layers exhibit incorrect cross-view correspondences—indicating that the quality of these correspondences directly governs the geometric consistency of the output.

Building on this insight, we explore whether geometric and motion consistency can be improved by directly supervising the correspondences in these dominant attention layers. Since the motion of dynamic objects can be described by the trajectories of physical points over time~\cite{doersch2023tapir}, we introduce \textbf{\ours}~(\textbf{M}ulti-\textbf{V}iew point \textbf{Track}ing \textbf{for} Novel-View \textbf{Gen}eration), a framework that leverages ground-truth multi-view point tracks as auxiliary supervision, where each track follows the same physical point within and across views. 

Specifically, we build a multi-view tracking head on top of local 4D correlation volumes computed from the query and key features of the selected attention layer, and jointly train it with the diffusion model. This encourages the model to encode motion-aware correspondences in its attention features, so that the target view more faithfully reflects the motion of the reference video. In addition, we introduce a multi-view correspondence loss that applies a cross-entropy objective directly to the attention map, encouraging each query token in the target view to attend to its corresponding ground-truth location. Together, we observe that these objectives improve both cross-view geometric consistency and intra-view temporal consistency of the generated novel-view videos.

To validate its generality, we apply our method to two camera-conditioning-only backbones, ReCamMaster~\cite{bai2025recammaster} and Redirector~\cite{park2025redirector}, and evaluate on the DAVIS~\cite{perazzi2016benchmark} and iPhone~\cite{gao2022monocular} benchmarks. Our method consistently improves both backbones, achieving the best scores on most VBench~\cite{huang2024vbench} visual-quality metrics while reaching state-of-the-art geometric consistency and camera accuracy comparable to both reconstruction-based and camera-conditioning-only baselines. Notably, it markedly improves geometric consistency and visual quality for dynamic objects, without requiring any explicit 3D reconstruction at inference time.
\section{Related Work}

\paragraph{Novel-View Generation Using Explicit 3D Representation.}
Recent video diffusion models~\cite{blattmann2023stable, yang2025cogvideox, wan2025wan} have demonstrated strong generation capability, ensuring visually plausible results. One line reconstructs explicit 3D representations from the input video and conditions a video diffusion model to synthesize newly visible regions from the target camera viewpoint.
Geometry-guided approaches~\cite{yu2025trajectorycrafter, ren2025gen3c, huang2025vivid4d, wang2025chronosobserver, zhao2026spatia, jeong2025reangle, kang2026egox, chen2025postcam, cao2025uni3c} first estimate intermediate geometric cues from the input video, such as depth maps, point cloud, or 3D representations, and use them as spatial conditions to guide the synthesis of disoccluded target-view regions. TrajectoryCrafter~\cite{yu2025trajectorycrafter} warps point clouds, GEN3C~\cite{ren2025gen3c} operates in latent space to generate novel views, and ChronosObserver~\cite{wang2025chronosobserver} synchronizes multi-view diffusion sampling through a hyperspace representation to produce time-synchronized, 3D-consistent multi-view videos. More recently, PostCam~\cite{chen2025postcam} fuses pose and visual signals through cross-attention shared by queries, while Infinite-Homography~\cite{kim2025infinite} conditions the generation of homography transformations as a lightweight geometric proxy that avoids explicit depth estimation.
Richer representations have also been explored, including pseudo-4D Gaussian fields from dense point tracking~\cite{bian2025gs} and feed-forward 4DGS reconstructors~\cite{yang2026neoverse}. Despite their strong spatial grounding, these approaches remain bottlenecked by the accuracy and completeness of off-the-shelf reconstruction models.

\paragraph{Novel-View Generation Using Camera Conditioning Only.}
An orthogonal line of work bypasses explicit geometry entirely, relying on camera pose conditioning and the generative prior of video diffusion models to implicitly reason about 3D structure~\cite{vanhoorick2024gcd, bai2025recammaster, bai2025syncammaster, park2025redirector, fan2025omniview, van2026anyview}.
GCD~\cite{vanhoorick2024gcd} is one of the earliest works to introduce camera-controlled dynamic novel-view generation by training a video generation model on Kubric~\cite{greff2022kubric}, a multi-view synthetic video dataset. To further improve generation quality, ReCamMaster~\cite{bai2025recammaster} is trained on a realistic synthetic dataset with more diverse camera trajectories. Redirector~\cite{park2025redirector} designs an additional camera rotational embedding for more accurate camera-controllable video generation.
A related line pursues joint space--time controllability: CAT4D~\cite{wu2025cat4d} trains a multi-view video diffusion model to synthesize novel-views at arbitrary camera poses and timestamps, enabling 4D reconstruction via deformable 3D Gaussians, while SpaceTimePilot~\cite{huang2025spacetimepilot} jointly conditions on camera trajectory and time to achieve generative rendering of dynamic scenes across both space and time.

\paragraph{Point Tracking.}
Tracking Any Point (TAP) formulates long-term pixel-level correspondence as a generalization of optical flow that explicitly handles occlusion~\cite{harley2022particle, doersch2022tap}. 
Early works such as PIPs~\cite{harley2022particle} and TAP-Net~\cite{doersch2022tap} suffered from occlusion fragility or per-frame independence, which TAPIR~\cite{doersch2023tapir} addressed by combining global matching-based initialization with a temporal refinement stage. CoTracker~\cite{karaev2024cotracker} further demonstrated that tracking points {jointly} via a transformer with cross-track attention substantially improves robustness under occlusion and fast motion, and CoTracker3~\cite{karaev2025cotracker3} unifies these ideas under a simpler architecture trained with pseudo-labels.

In the multi-view setting, MV-TAP~\cite{koo2026mv} aggregates spatio-temporal information across views via cross-view attention for robust 2D trajectory estimation, while MVTracker~\cite{rajivc2025mvtracker} targets feed-forward 3D point tracking by fusing multi-view features. These formulations yield geometrically consistent correspondences beyond what monocular trackers can recover, motivating their use as an auxiliary supervision signal to strengthen the geometric feature learning of video diffusion models.

\section{Preliminaries: Video Diffusion Transformer for Novel-View Generation}


Here, we explain the details of camera-conditioning-only frameworks for novel-view video generation, which aim to synthesize a video from a novel camera viewpoint given a monocular reference video. A typical video generation model for novel-view synthesis consists of two main components: VAE and  Diffusion Transformer (DiT). Given a reference video $X_\text{ref} \in \mathbb{R}^{F \times H \times W \times 3}$ where $F$, $H$, $W$, and $3$ denote the number of frames, spatial height, width, and RGB channels, respectively, the VAE encodes it into a latent representation $z_\text{ref} \in \mathbb{R}^{f \times h \times w \times d_{\text{video}}}$ where $(f, h, w, d_{\text{video}})$ are the temporal, spatial, and channel dimensions in the latent space. The DiT then denoises a target-view latent $z_\text{tgt} \in \mathbb{R}^{f \times h \times w \times d_{\text{video}}}$ conditioned on $z_\text{ref}$ to generate the novel-view video.

Specifically, the VAE downsamples the input by $4\times$ temporally and $16\times$ spatially. The DiT $\mathbf{v}_\theta$ takes the concatenation of $z_\text{ref}$ and a noisy $z_\text{tgt}$ as input, and predict the velocity field: 
\begin{equation}
\hat{\mathbf{v}} =  \mathbf{v}_\theta\!\left([\, z_\text{ref},\, z_\text{tgt} \,], \, t,\, c,\, \mathrm{cam}_\text{tgt} \right),
\end{equation}
where $[\,\cdot\,,\,\cdot\,]$ denotes concatenation along the token dimension, $t$ is the flow matching timestep, $c$ is the text caption, and $\mathrm{cam}_\text{tgt}$ is the target camera trajectory. The model is trained via flow matching~\cite{lipman2023flowmatchinggenerativemodeling}.

\begin{wrapfigure}{rt}{0.35\textwidth}
    \vspace{-15pt}
    \centering
    \includegraphics[width=\linewidth]{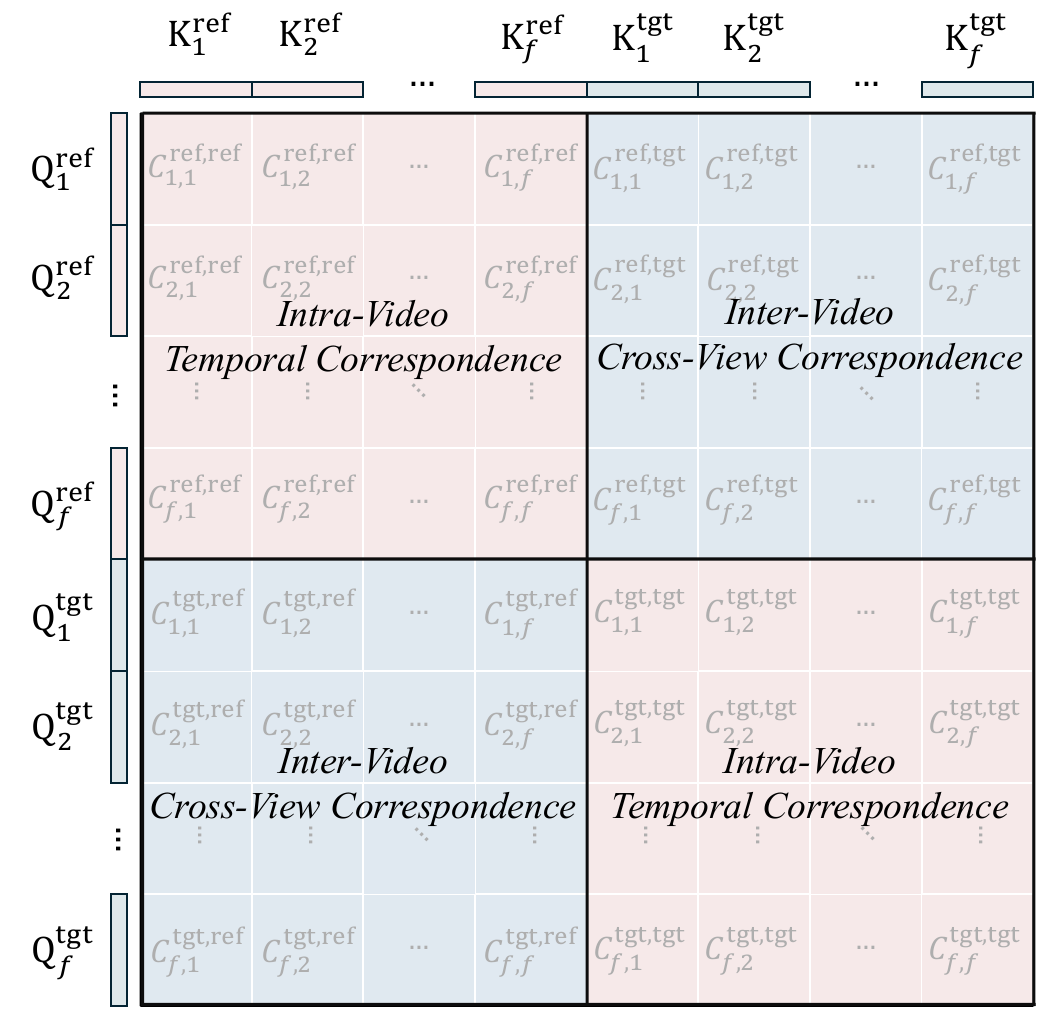}
    \caption{\textbf{Illustration of full 3D attention in video DiT for novel-view generation.} 
    The attention jointly captures intra-video temporal and inter-video cross-view correspondences.}
    \vspace{-15pt}
    \label{fig:attention_map}
\end{wrapfigure}

\paragraph{3D Attention Across Reference and Target Latent Tokens.}
Since $z_\text{ref}$ and $z_\text{tgt}$ are concatenated as input to DiT, they jointly participate in 3D attention. At each transformer layer $l$ and flow matching timestep $t$, the $i$-th latent frame ($i \in \{1, \dots, f\}$) of each view, either reference or target, is projected into query and key matrices:
\begin{equation}
Q^\text{ref}_{i},\, K^\text{ref}_{i},\, Q^\text{tgt}_{i},\, K^\text{tgt}_{i} \in \mathbb{R}^{hw \times d_{\text{head}}},
\end{equation}
where $d_{\text{head}}$ is the per-head channel dimension and we omit the attention head for brevity.
The attention weight matrix $\mathcal{C}^{v_1,v_2}_{i,j}$, where each entry measures the similarity between the $i$-th frame of view $v_1$ and the $j$-th frame in view $v_2$, is computed as:

\begin{equation}
\label{eq:matching_cost}
\mathcal{C}^{v_1,v_2}_{i,j} = \mathrm{Softmax}\!\left(\frac{Q^{v_1}_{i} \bigl(K^{v_2}_{j}\bigr)^\top}{\sqrt{d_{\text{head}}}}\right),
\quad v_1, v_2 \in \{\text{ref},\, \text{tgt}\}.
\end{equation}

Fig.~\ref{fig:attention_map} illustrates the resulting attention map structure.

\section{Analysis}
\label{sec:analysis}
Since camera-conditioning-only models~\cite{bai2025recammaster,park2025redirector} implicitly exchange information through 3D attention both across the reference and target views and over time within each view, we analyze their attention maps to examine how geometry and motion are encoded inside the models.
Following~\cite{nam2025emergenttemporalcorrespondencesvideo}, we conduct our main analysis on ReCamMaster~\cite{bai2025recammaster}, and show that its 3D attention maps contain emergent token-level correspondences across layers and denoising timesteps.
By extracting matches from attention weights and comparing them with pseudo ground-truth point tracks, we identify three types of emergent correspondences: \emph{intra-video temporal correspondences} in the reference view, \emph{intra-video temporal correspondences} in the target view, and \emph{inter-video cross-view correspondences}. The intra-video correspondences capture motion-related temporal consistency within each view, whereas the inter-video correspondences are responsible for geometric alignment across views.
Our analysis is also applicable to other camera-conditioning-only architectures, as demonstrated by our analysis of Redirector~\cite{park2025redirector} in Appendix~\ref{sec:redirector_analysis}.

\begin{wrapfigure}{r}{0.45\textwidth}
    \vspace{-30pt}
    \centering
    \includegraphics[width=\linewidth]{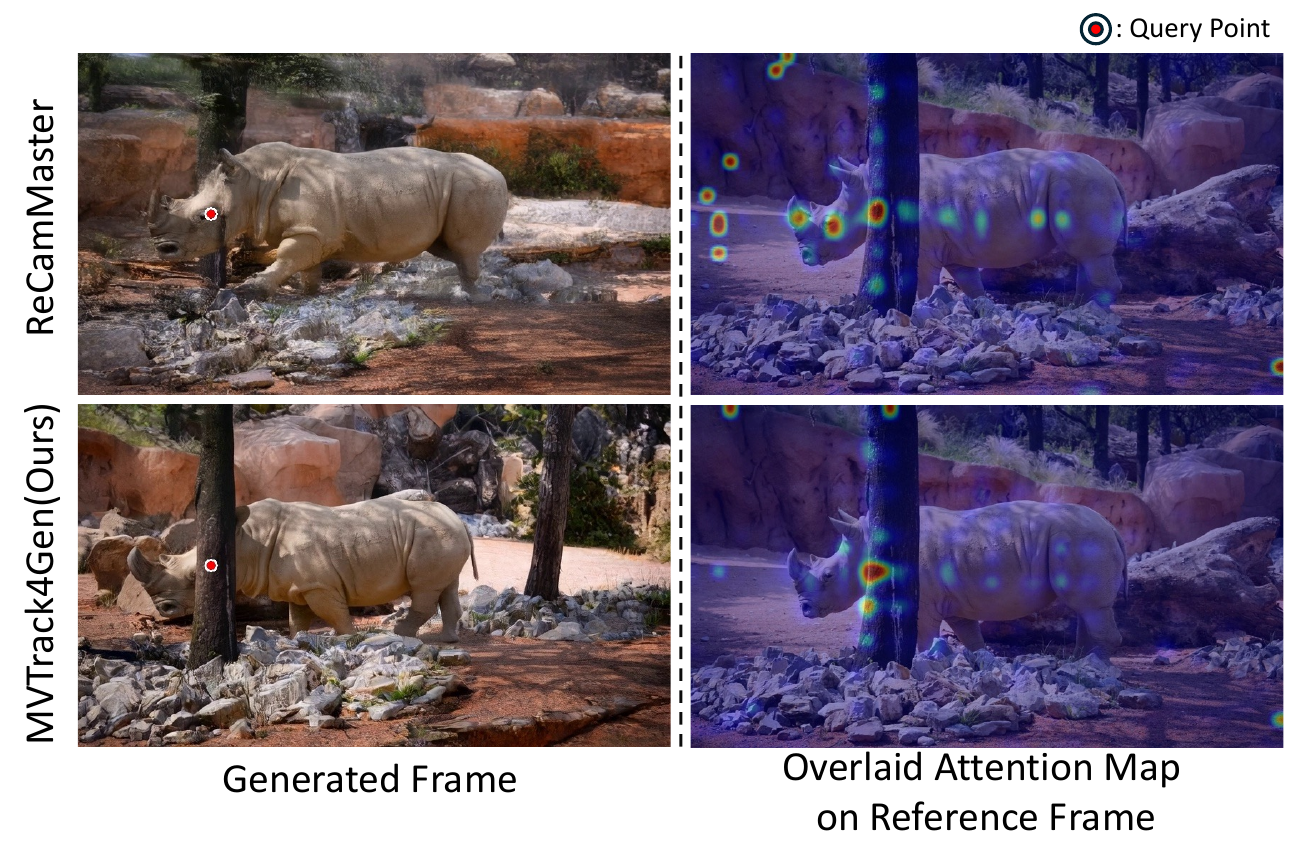}
    \vspace{-20pt}
     \caption{\textbf{Cross-View Attention Visualization.}
    For the same query point in the generated frame, ReCamMaster attends to incorrect regions in the reference frame, whereas \ours\ localizes attention on the corresponding object, enabling more consistent motion across views.}
    \vspace{-20pt}
    \label{fig:attention_map_vis}
\end{wrapfigure}



\subsection{Analysis Setup}
We conduct our analysis on the MultiCamVideo dataset~\cite{bai2025recammaster}, which provides time-synchronized multi-view recordings of dynamic scenes with ground-truth camera trajectories, naturally yielding paired reference--target videos. We sample 40 scenes and randomly select two views per scene as the reference and target.

To obtain pseudo ground-truth correspondences that span both temporal and cross-view axes, we use MV-TAP~\cite{koo2026mv}, an off-the-shelf multi-view point tracker that jointly tracks points across reference and target video frames. This produces multi-view pseudo ground-truth point tracks $\mathcal{T} = \{ p^{v, \text{GT}}_{i} \}$ with visibility $\mathcal{O} = \{ o^{v,\text{GT}}_{i} \}$, 
where $v$ and $i$ index the view and frame, respectively. More details are in Appendix~\ref{sec:analysis_eval_details}.

\subsection{Evaluating Attention-based Correspondences}
\label{sec:corresp_3d_attention}

\paragraph{Correspondence in 3D attention map.}
We investigate token-level correspondences from the 3D attention map defined in Eq.~\ref{eq:matching_cost}.
For a pair of latent frames, the attention matrix $C^{v_1,v_2}_{i,j}$ measures how each query features in the $i$-th frame of view $v_1$ attends to key features in the $j$-th frame of view $v_2$, where $v_1,v_2 \in \{{\mathrm{ref}, \mathrm{tgt}}\}$.
Given a query feature in the $i$-th frame of view $v_1$, we take the key feature with the highest attention weight in the $j$-th frame of view $v_2$ as its forward match.

To enable a more accurate analysis, we apply a cycle-consistency check to filter out unreliable matches.
A forward match is accepted as a \emph{reliable correspondence} only if matching backward from the destination latent frame to the source latent frame returns to the original query token.
Formal definitions are provided in Appendix~\ref{sec:correspondence_details}.


\paragraph{Matching accuracy and harmonic mean.}
We measure matching accuracy using the Percentage of Correct Keypoints (PCK), which counts a query point as a positive only if (i) it is reliable under the cycle-consistency check and (ii) its forward match coincides with the corresponding co-visible point in the pseudo ground-truth $\mathcal{T}$ on the latent grid. PCK is reported separately for \emph{intra-video temporal correspondences} in the reference and target views, and for \emph{inter-video cross-view correspondences}, averaged over all query points across latent frames and scenes.
Beyond matching accuracy alone, we further analyze two attention-weight-based quantities to characterize not only whether the selected match is correct, but also how the attention distribution supports information exchange across layers and denoising timesteps: an attention score, which measures how much attention weight is assigned to the correct match, and a confidence score, which measures how sharply the attention distribution is localized.
We then compute the harmonic mean of three normalized metrics—matching accuracy, attention score, and confidence score—to identify layers in which all three metrics are simultaneously high, thereby assessing how the correspondences emerging inside the attention layers contribute to generation; further details are provided in Appendix~\ref{sec:correspondence_details}.



\subsection{Results}
\begin{figure}[t]
    \centering
    \includegraphics[width=0.9\linewidth]{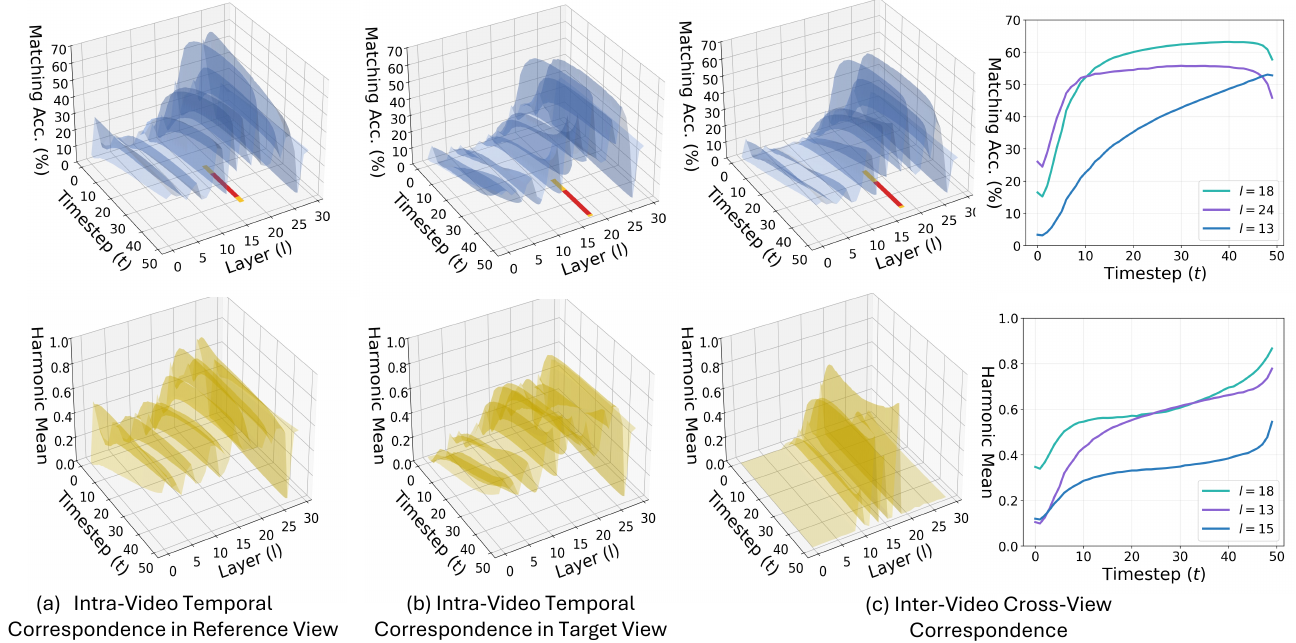}
    \caption{\textbf{Matching Accuracy and Harmonic Mean in ReCamMaster~\cite{bai2025recammaster}.} We visualize matching accuracy (top) and the harmonic mean of matching accuracy, attention score, and confidence score (bottom) across diffusion layers and denoising timesteps. Results are shown for intra-video temporal correspondence in the reference and target views in (a) and (b), respectively, and for inter-video cross-view correspondence in (c). The rightmost column plots the top three layers with the highest cross-view correspondence scores. This figure shows that accurate cross-view and temporal matching emerge at specific intermediate layers.}
    \label{fig:analysis}\vspace{-10pt}
\end{figure}

Fig.~\ref{fig:analysis} shows the matching accuracy and harmonic mean by comparing correspondences from the attention maps and the pseudo ground-truth matches $\mathcal{T}$ across denoising timesteps and layers. We observe correspondence cues both within each view over time and across the reference and target views, which we refer to as \emph{intra-video temporal correspondences} and \emph{inter-video cross-view correspondences}, respectively.

We observe that specific intermediate layers exhibit strong matching performance for both intra-video temporal correspondences and inter-video cross-view correspondences, indicating that the same layer range supports motion tracking within each view and geometric alignment across views, as visualized in Fig.~\ref{fig:analysis}~(c), with a dominant peak around the 18th diffusion layer. The harmonic mean follows the same layer--timestep trend as the matching accuracy, suggesting that cross-view matching capability is encapsulated within only specific layers. Moreover, in failure cases of generation as visualized in Fig.~\ref{fig:attention_map_vis}, the attention map at the 18th layer fails to attend to the corresponding regions in the reference video. Together, these results motivate our design choice of explicitly reusing features from this layer range as a unified correspondence signal for both temporal and cross-view consistency.

\section{Methodology}
\begin{figure}[t]
    \centering
    \vspace{-10pt}
    \includegraphics[width=\linewidth]{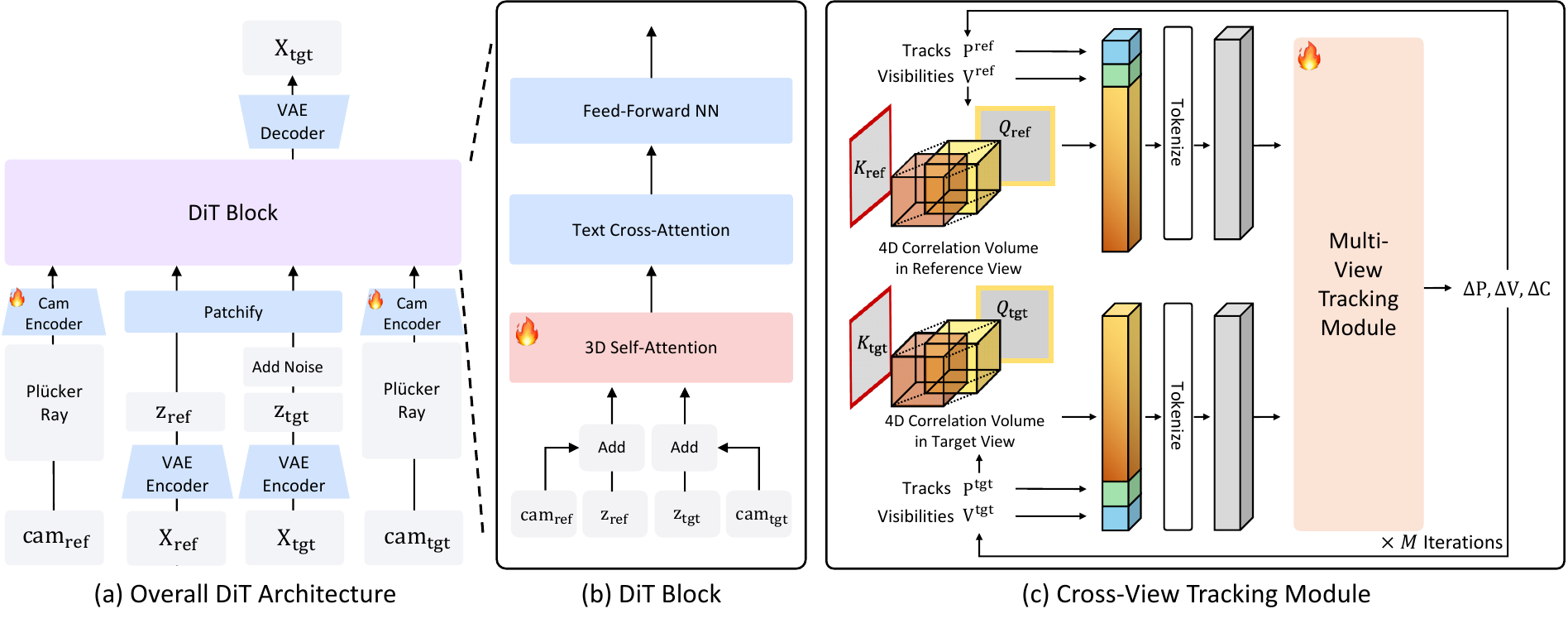}
    \vspace{-15pt}
    \caption{\textbf{Main architecture.}
    We jointly train a camera-controlled DiT and a multi-view tracking module that shares query and key features from the input of the DiT's 3D self-attention layers.
    From these shared features, the tracking module constructs \emph{intra-video temporal correlation} for temporal consistency and \emph{inter-video cross-view correlation} for geometric correspondence.}
    \vspace{-10pt}
    \label{fig:main_architecture}
\end{figure}

Given a reference video $X_\text{ref}$ captured from a known camera trajectory $\mathrm{cam}_\text{ref}$ and a target camera trajectory $\mathrm{cam}_\text{tgt}$, our goal is to synthesize a geometrically consistent target video $X_\text{tgt}$ that faithfully depicts the same dynamic scene from the new viewpoint. We achieve this by jointly training a camera-controlled video diffusion model and a multi-view tracking module that shares the query and key features from the diffusion model's 3D attention layers. We further directly supervise the 3D attention maps with a multi-view correspondence loss, guiding each query feature to attend to the correct corresponding region across views and time. Together, these objectives provide geometric and motion supervision to the generation process. An overview of our framework is illustrated in Fig.~\ref{fig:main_architecture}.

\subsection{Improved Camera Encoding}
We adopt ReCamMaster~\cite{bai2025recammaster} and Redirector~\cite{park2025redirector} as our backbones, but condition on both the extrinsics and intrinsics of the reference and target views, rather than the target extrinsics alone. This richer conditioning improves geometric consistency in novel-view generation. Each camera is encoded as a dense Pl\"ucker ray map~\cite{sitzmann2021light}, in which every pixel is represented by a 6D ray derived from the extrinsics and intrinsics. Pl\"ucker maps from both views are injected into each DiT layer; additional details are provided in Appendix~\ref{sec:camera_details}.

\subsection{Multi-View Point Tracking as Geometric Supervision}
\label{sec:point_tracking}

\paragraph{Multi-Scale 4D Correlation Volume in Each View.}
We construct a multi-scale local 4D correlation volume~\cite{cho2024local} in each view, leveraging the query--key similarity in the 3D attention map. This is motivated by our analysis in Sec.~\ref{sec:analysis}, which shows that the \emph{intra-video temporal correspondences} in the 3D attention map inherently encode motion across time. 
Local query and key features $q^{v}_{i}$ and $k^{v}_{j}$ are bilinearly sampled within a $\Delta$-sized neighborhood centered at the query points $p^{v}_i=(x^{v}_i,y^{v}_i)$ of the $i$-th frame in each view $v \in \{\text{ref, tgt}\}$ and the estimated points $\hat{p}^{v}_{j} = (\hat{x}^{v}_{j}, \hat{y}^{v}_{j})$ of the $j$-th frame. The local 4D correlation volume is then computed via softmax:
\begin{equation}
    \label{eq:local_cost}
    \operatorname{Corr}^{v}_{i,j} = \mathrm{Softmax}\!\left(\frac{q^{v}_{i} \bigl(k^{v}_{j}\bigr)^{\top}}{\sqrt{d_{\text{head}}}}\right) \in \mathbb{R}^{(2\Delta+1)^4},\quad v \in \{\text{ref},\, \text{tgt}\}.
\end{equation}
Each temporal correlation volume within each view, $\operatorname{Corr}^{\text{ref}}$ and $\operatorname{Corr}^{\text{tgt}}$, encodes the motion within its corresponding view and serves as the input to the multi-view tracking module. Query points are sampled among all video frames, since any point should be trackable and its motion detectable wherever it newly appears in the video. This encourages the query feature to be more similar to its corresponding ground-truth key feature, which we empirically find makes the generated motion more faithful to the reference. More details are described in Appendix~\ref{sec:hierarchical_local_4d_correlation}.

\paragraph{Multi-View Point Tracking Head.}
Built on the correlation volumes described above, we adopt a transformer-based multi-view tracking head following~\cite{koo2026mv}. At each iteration, the tracking head $\Psi$, equipped with factorized temporal and multi-view attention, predicts residuals $(\Delta P, \Delta V, \Delta C) = \Psi(G)$, where input tokens $G$ are formed by concatenating current visibility $V$, confidence $C$, and local 4D correlation volume $\operatorname{Corr}^{v}_{i,j}$ in each view. Visibility and confidence are initialized to zero for both views, i.e., $V = C = 0$.

\subsection{Training Objectives}
\label{sec:training}
We supervise the  correspondences in the 3D attention maps with two additional objectives: a multi-view point-tracking loss $\mathcal{L}_{\text{track}}$ and a multi-view correspondence loss $\mathcal{L}_{\text{corr}}$. The tracking objective makes the model follow the dynamic motion of the reference video by jointly training the diffusion model with the multi-view point tracking head. The multi-view correspondence loss enforces geometric consistency—not only between the reference view and the target view, but also within the target view itself—by strengthening matching information directly into the 3D attention map to guide where the model should attend.

The total loss is $\mathcal{L}_{\text{total}} = \mathcal{L}_{\text{diff}} + \lambda_{\text{track}}\,\mathcal{L}_{\text{track}} + \lambda_{\text{corr}}\,\mathcal{L}_{\text{corr}}$, where $\lambda_{\text{track}} = \lambda_{\text{corr}} = 0.01$.
The diffusion loss follows the rectified-flow formulation, $\mathcal{L}_{\text{diff}} = w(t)\,\bigl\| \mathbf{v}_\theta - (\epsilon - x_0) \bigr\|^2_{\text{tgt}}$, where $w(t)$ is a timestep-dependent weighting and the squared error is evaluated only on the target-view tokens. The tracking loss jointly supervises the multi-view point-tracking head and diffusion model with ground-truth trajectories and decomposes into $\mathcal{L}_{\text{track}} = \lambda_{\text{seq}}\,\mathcal{L}_{\text{seq}} + \lambda_{\text{conf}}\,\mathcal{L}_{\text{conf}} + \lambda_{\text{vis}}\,\mathcal{L}_{\text{vis}}$, where $\mathcal{L}_{\text{seq}}$ is a visibility-weighted Huber loss on the predicted 2D point coordinates, $\mathcal{L}_{\text{conf}}$ is a probabilistic confidence loss that calibrates the predicted per-point confidence against the regression error, and $\mathcal{L}_{\text{vis}}$ is a binary cross-entropy on the predicted visibility logits.

The multi-view correspondence loss $\mathcal{L}_{\text{corr}}$ directly supervises the attention weight matrix defined in Eq.~\ref{eq:matching_cost} with a cross-entropy objective. For each query point, the ground-truth multi-view point tracks provide its corresponding locations and visibility across all frames and views. Since each frame contains a single ground-truth correspondence for a given query point when the target point is visible, we formulate this as a single-label classification problem. Specifically, we supervise the attention weight matrix to assign the highest probability to the matching token in each visible reference or target frame. More details are provided in Appendix~\ref{sec:supp_loss}.

\section{Experiments}

\begin{table}[t]
    \centering
    \renewcommand{\arraystretch}{0.9}
    \caption{\textbf{Quantitative comparison on the DAVIS dataset~\cite{perazzi2016benchmark}.}
    The best score for each metric is in \textbf{bold}, and the colored numbers denote the change relative to each backbone, with \gain{green} indicating a gain and \loss{red} a loss.}
    \label{tab:davis_results}
    \setlength{\tabcolsep}{3pt}
    \footnotesize
    \resizebox{\linewidth}{!}{
    \begin{tabular}{l|cccccc|cc|ccc}
    \toprule
    \multirow{2}{*}{Method} & \multicolumn{6}{c|}{Visual Quality$\uparrow$} & \multicolumn{2}{c|}{Geo. Consist.} & \multicolumn{3}{c}{Camera Accuracy} \\
    & \makecell{Subject\\Consistency} & \makecell{Background\\Consistency} & \makecell{Aesthetic\\Quality} & \makecell{Imaging\\Quality} & \makecell{Temporal\\Flickering} & \makecell{Motion\\Smoothness} & MEt3R$\downarrow$ & MEt3R$_{\text{dynamic}}\downarrow$& mRotErr ($^\circ$)$\downarrow$ & mTransErr$\downarrow$ & mCamMC$\downarrow$ \\
    \midrule
    \textit{Explicit 3D Lifting} \\
    \quad GEN3C~\cite{ren2025gen3c}                        & 0.856 & 0.894 & 0.461 & 0.582 & 0.950 & {0.980} & {0.290} & 0.328 & 2.538 & 0.127 & 0.163 \\
    \quad TrajectoryCrafter~\cite{yu2025trajectorycrafter} & 0.847 & 0.880 & 0.438 & 0.550 & 0.942 & 0.970 & 0.291 & 0.306 & 10.126 & 0.190 & 0.355 \\
    \quad CogNVS~\cite{chen2025reconstruct}                & 0.811 & 0.862 & 0.417 & 0.530 & {0.959} & 0.978 & 0.333 & 0.346 & 10.439 & 0.228 & 0.400 \\
    \quad NeoVerse~\cite{yang2026neoverse}                 & 0.858 & 0.878 & 0.446 & 0.591 & 0.954 & 0.983 & {0.302} & 0.323 & 4.705 & 0.159 & 0.228 \\
    \midrule
    \textit{Camera Conditioning Only} \\
    \quad ReCamMaster~\cite{bai2025recammaster}            & 0.904 & 0.907 & 0.502 & 0.652 & \textbf{0.962} & 0.985 & 0.337 & 0.369 & 3.660 & 0.113 & 0.169 \\
    \quad Redirector~\cite{park2025redirector}             & 0.897 & 0.911 & {0.506} & {0.680} & 0.952 & {0.985} & 0.318 & 0.395 & \textbf{1.714} & 0.086 & 0.109 \\
   \midrule
   \textbf{\ours~\textsubscript{ReCamMaster} } & 0.892\,\loss{-.012} & 0.909\,\gain{+.002} & 0.507\,\gain{+.005} & 0.685\,\gain{+.033} & 0.953\,\loss{-.009} & 0.984\,\loss{-.001} & {0.274}\,\gain{-.063} & \textbf{0.287}\,\gain{-0.082} & 1.858\,\gain{-1.802} & 0.100\,\gain{-.013} & 0.125\,\gain{-.044} \\
    \textbf{\ours~\textsubscript{Redirector} }   & \textbf{0.905}\,\gain{+.008} & \textbf{0.919}\,\gain{+.008} & \textbf{0.508}\,\gain{+.002} & \textbf{0.687}\,\gain{+.007} & 0.956\,\gain{+.004} & \textbf{0.986}\,\gain{+.001} & \textbf{0.267}\,\gain{-.051} & 0.349\,\gain{-0.036} & 1.718\,\loss{+.004} & \textbf{0.073}\,\gain{-.013} & \textbf{0.097}\,\gain{-.012} \\
    \bottomrule
    \end{tabular}
    }
    \vspace{-10pt}
\end{table}

\begin{wraptable}[11]{R}{0.46\linewidth}
    \centering
    \vspace{-22pt}
    \caption{\textbf{Quantitative comparison on the iPhone~\cite{gao2022monocular} dataset.}}
    \label{tab:iphone_results}
    \vspace{7pt}
    \setlength{\tabcolsep}{3pt}
    \footnotesize
    \resizebox{\linewidth}{!}{
    \begin{tabular}{l|ccc|c}
    \toprule
    \multirow{2}{*}{Method} & \multicolumn{3}{c|}{Visual Quality} & Geo. Consist. \\
    & PSNR$\uparrow$ & SSIM$\uparrow$ & LPIPS$\downarrow$ & MEt3R$\downarrow$ \\
    \midrule
    \textit{Explicit 3D Lifting} \\
    \quad GEN3C~\cite{ren2025gen3c}                        & 10.419 & 0.270 & 0.699 & 0.382 \\
    \quad TrajectoryCrafter~\cite{yu2025trajectorycrafter} & 10.271 & 0.276 & 0.766 & 0.370 \\
    \quad CogNVS~\cite{chen2025reconstruct}                & 10.105 & 0.280 & 0.774 & 0.406 \\
    \quad NeoVerse~\cite{yang2026neoverse}                 & {11.886} & {0.394} & {0.579} & 0.493 \\
    \midrule
    \textit{Camera Conditioning Only} \\
    \quad ReCamMaster~\cite{bai2025recammaster}            & 11.005 & {0.338} & 0.705 & 0.461 \\
    \quad Redirector~\cite{park2025redirector}             & 11.447 & 0.329 & 0.720 & 0.580 \\
    \midrule
        {\textbf{\ours~\textsubscript{ReCamMaster}}}       
    & 11.521\,\gain{+.516} 
    & 0.270\,\loss{-.068} 
    & 0.640\,\gain{-.065} 
    & 0.381\,\gain{-.080} \\

    {\textbf{\ours~\textsubscript{Redirector}}}       
    & 11.830\,\gain{+.383} 
    & 0.283\,\loss{-.046} 
    & 0.638\,\gain{-.082} 
    & 0.397\,\gain{-.183} \\
    \bottomrule
    \end{tabular}
    \vspace{-15pt}
    }
\end{wraptable}

\subsection{Experimental Setup}
\paragraph{Implementation Details.}
We build on ReCamMaster~\cite{bai2025recammaster} and Redirector~\cite{park2025redirector} as our backbones, fine-tuning the 3D attention layers and camera encoder while keeping all other parameters frozen. The multi-view point tracking head estimates point tracks using query-key features from the 18th layer of the DiT.
We train on 4 NVIDIA H100 GPUs using mixed-precision training and the AdamW optimizer with a learning rate of $1{\times}10^{-4}$ for 13,000 iterations with a batch size of 16. All videos are used at $480{\times}832$ resolution with 81 frames for training.

\paragraph{Datasets.}
We train our model on the combined Kubric and MultiCamVideo described in Sec.~\ref{sec:training_data}.
For evaluation, we adopt two benchmarks covering complementary regimes: (i) the DAVIS dataset~\cite{perazzi2016benchmark} for in-the-wild monocular videos with diverse object and camera motion, and (ii) the iPhone dataset~\cite{gao2022monocular} for casual hand-held captures of dynamic scenes. 


\paragraph{Baselines.}
We compare against two families of camera-controlled video generation methods.
\emph{Explicit 3D Lifting} approaches first reconstruct geometry and re-render under the target camera; in this category we include GEN3C~\cite{ren2025gen3c}, TrajectoryCrafter~\cite{yu2025trajectorycrafter}, CogNVS~\cite{chen2025reconstruct}, and NeoVerse~\cite{yang2026neoverse}.
\emph{Camera Conditioning Only} approaches rely solely on the diffusion prior together with camera conditioning, without explicit 3D reconstruction; here we compare with ReCamMaster~\cite{bai2025recammaster} and Redirector~\cite{park2025redirector}.

\paragraph{Evaluation Metrics.}
We evaluate along three complementary axes: generation quality, geometric consistency, and camera accuracy.
For generation quality, we report PSNR, SSIM, and LPIPS against ground-truth target views on the iPhone dataset~\cite{gao2022monocular}; on DAVIS, where ground-truth views are unavailable, we follow VBench~\cite{huang2024vbench} and report Subject Consistency, Background Consistency, Aesthetic Quality, Imaging Quality, Temporal Flickering, and Motion Smoothness.
For geometric consistency, we report MEt3R~\cite{asim2025met3r}, which measures multi-view geometric consistency via feed-forward 3D reconstruction, and additionally report MEt3R$_\text{dynamic}$, which restricts this measure to dynamic regions.
For camera accuracy, evaluated on DAVIS, we estimate camera trajectories from the generated videos using Depth Anything3~\cite{lin2025depth} and report mean rotation error (mRotErr), mean translation error (mTransErr), and mean camera-motion consistency (mCamMC) against the target trajectory.

\begin{figure}[t]
    \centering
    \includegraphics[width=\linewidth]{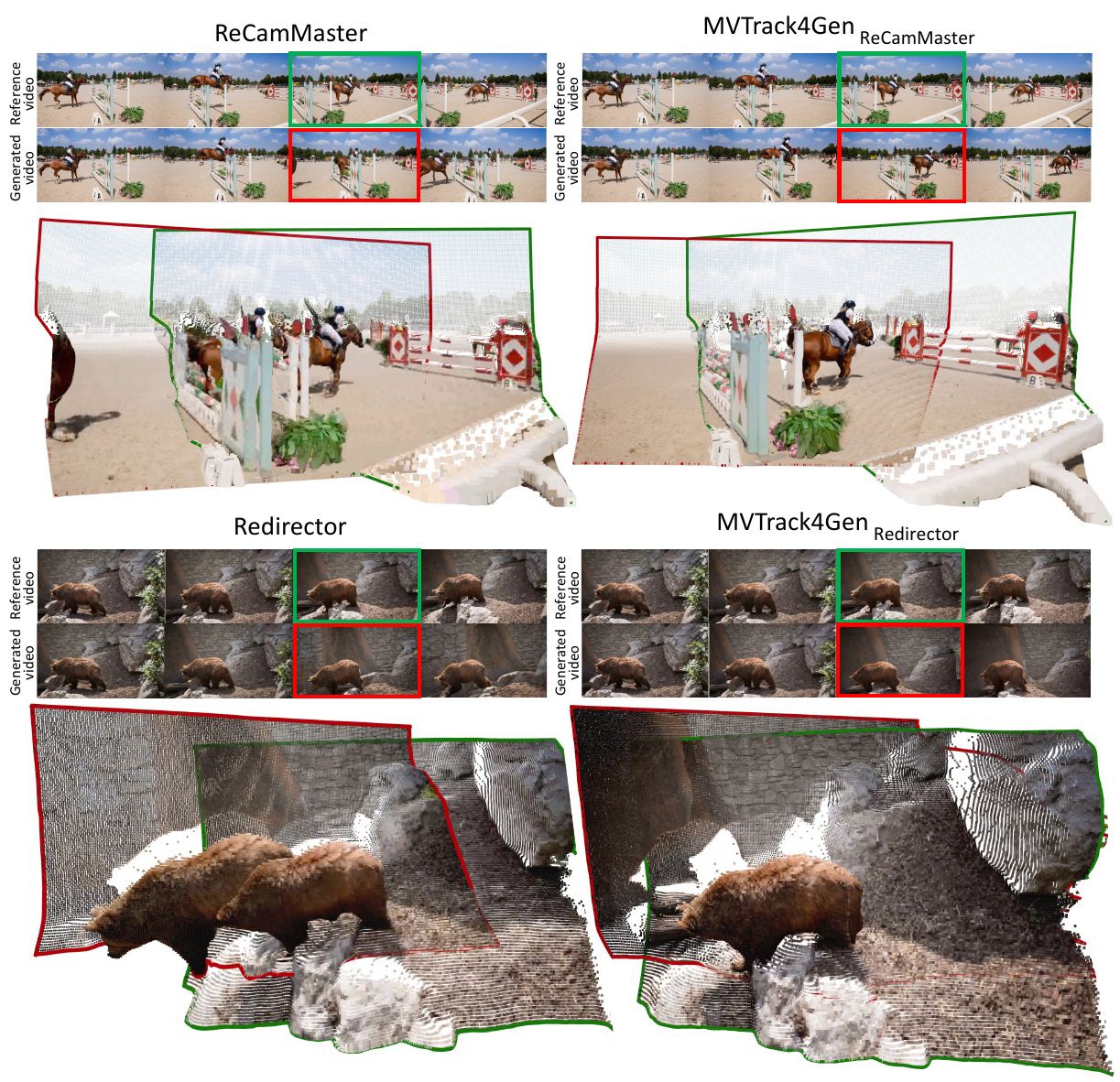}
    \vspace{-15pt}
    \caption{\textbf{Qualitative results on the DAVIS dataset~\cite{perazzi2016benchmark}.}
    We attach our \ours\ to two backbones, ReCamMaster~\cite{bai2025recammaster} and Redirector~\cite{park2025redirector}, and compare against each baseline.
    For every example, we show the reference and generated frames, together with their point-cloud renderings reconstructed by Depth Anything3~\cite{lin2025depth} for the \textcolor{refgreen}{\textbf{reference view}} and \textcolor{genred}{\textbf{target view}}.
    By strengthening cross-view correspondences, \ours\ generates target frames that remain faithfully aligned with the reference view, even for dynamic objects, while preserving scene structure, appearance, and coherent motion across views.}
    \label{fig:davis_main_qual}\vspace{-10pt}
\end{figure}

\subsection{Quantitative Results}
We compare \ours\ against state-of-the-art baselines from both the \textit{Explicit 3D Lifting} and \textit{Camera Conditioning Only} paradigms on the DAVIS~\cite{perazzi2016benchmark} and iPhone~\cite{gao2022monocular} datasets, as summarized in Tab.~\ref{tab:davis_results} and Tab.~\ref{tab:iphone_results}. On DAVIS, \ours\ achieves the best overall performance across all three evaluation axes. For visual quality, it ranks first on Subject Consistency, Background Consistency, Aesthetic Quality, Imaging Quality, and Motion Smoothness. For geometric consistency, it attains the lowest MEt3R and MEt3R$_\text{dynamic}$, outperforming all baselines. For camera accuracy, it achieves the lowest mTransErr and mCamMC while remaining competitive on mRotErr. On iPhone, \ours\ consistently outperforms all \textit{Camera Conditioning Only} baselines on PSNR, LPIPS, and MEt3R, demonstrating that strengthening cross-view correspondence is highly effective even without any explicit 3D representation. Although \textit{Explicit 3D Lifting} methods attain higher pixel-level fidelity on iPhone owing to their direct reliance on reconstructed geometry, \ours\ achieves comparable or even better geometric consistency without using any explicit 3D structure at inference. This confirms that strengthening cross-view correspondence in a video diffusion model effectively bridges the gap between the two paradigms.

\begin{figure}[t]
    \centering
    \includegraphics[width=\linewidth]{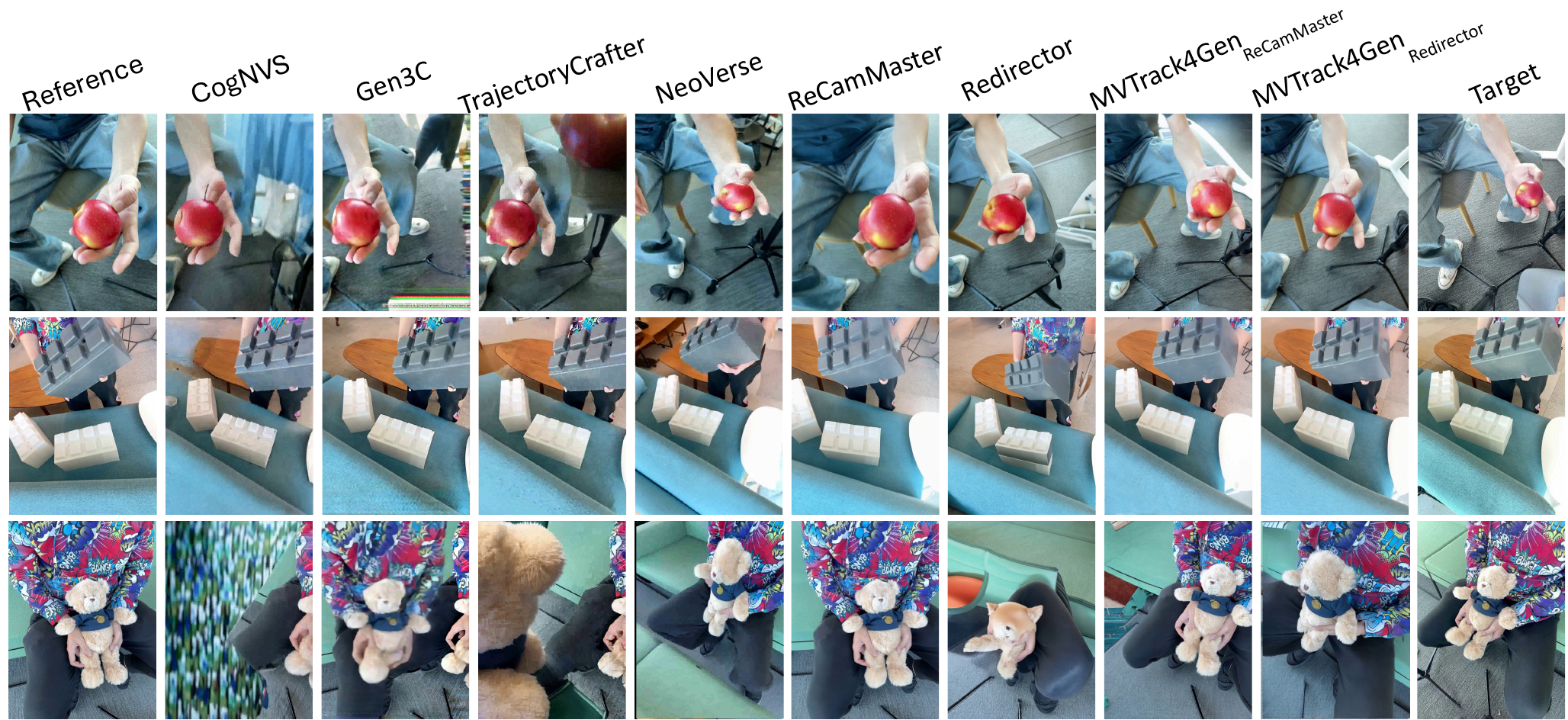}
    \caption{\textbf{Qualitative results on the iPhone dataset~\cite{gao2022monocular}.}
    We visualize novel-views generated from the reference video by each method, alongside the ground-truth target view. Our \ours, built on both ReCamMaster~\cite{bai2025recammaster} and Redirector~\cite{park2025redirector}, produces geometrically consistent novel-views that faithfully preserve the scene structure and appearance of the reference, while prior methods exhibit viewpoint inaccuracies, geometric distortions, or texture degradation.}
    \label{fig:iphone_main_qual}
\end{figure}

\subsection{Qualitative Results}
Fig.~\ref{fig:davis_main_qual} presents qualitative comparisons on the DAVIS dataset, where we attach \ours\ to both ReCamMaster~\cite{bai2025recammaster} and Redirector~\cite{park2025redirector} backbones and show, for each method, the generated frames together with their point-cloud renderings reconstructed by DepthAnything3~\cite{lin2025depth} for the reference and target views. The baselines produce plausible-looking frames but lack geometric grounding under viewpoint change, particularly on dynamic objects: their reference and generated point clouds fail to align on the moving foreground, indicating inconsistent scene geometry across views. In contrast, \ours\ produces point clouds that align faithfully between the reference and target views even for dynamic objects, preserving scene structure and keeping the motion of moving foreground objects coherent across viewpoints.

Fig.~\ref{fig:iphone_main_qual} further presents comparisons on challenging real-world scenes from the iPhone dataset. Camera Conditioning Only baselines such as ReCamMaster and Redirector synthesize plausible frames but exhibit geometric inconsistencies under viewpoint change, with foreground objects appearing at incorrect depths or undergoing non-rigid warping. Explicit 3D Lifting methods such as CogNVS, GEN3C, and TrajectoryCrafter preserve coarse scene structure but introduce noticeable artifacts in regions where depth estimation is unreliable, including severe distortions of dynamic foreground objects. In contrast, \ours\ synthesizes geometrically consistent novel-views that faithfully preserve both scene structure and appearance from the reference, even under significant viewpoint changes and complex object motion. These observations are consistent with our quantitative gains in geometric consistency (MEt3R) and camera accuracy, confirming that strengthening cross-view correspondence in a video diffusion model leads to both visually and geometrically faithful novel-view synthesis. More generation results are provided in Sec.~\ref{sec:more_gen_qual}.

\subsection{Ablation Studies}
We conduct ablation studies on the iPhone dataset~\cite{gao2022monocular}, progressively adding components to the ReCamMaster baseline; results are summarized in Tab.~\ref{tab:arch_ablation}. All variants are fine-tuned for 10k iterations from ReCamMaster, with ``+'' indicating additions to (i).
Configuration (ii) applies our multi-view correspondence loss on the 3D attention map. This attention-level supervision substantially improves geometric consistency and camera accuracy over (i) while leaving image quality largely unchanged, showing that supervising \emph{where each query attends} strengthens cross-view geometric alignment.
Configuration (iii) instead adds our multi-view tracking module, where query points are tracked within each view. By making the model follow the reference motion, it improves motion fidelity and generation quality.
Configuration (iv) combines both: the tracking module improves motion fidelity while the correspondence loss enforces geometric consistency and camera accuracy. These two components are complementary, attaining the best visual quality, geometric consistency, and camera accuracy across all axes.

\begin{table}[t]
\centering
\caption{\textbf{Architecture ablation on the iPhone dataset~\cite{gao2022monocular}.} We progressively add components to the ReCamMaster baseline to validate each design choice.}
\label{tab:arch_ablation}
\resizebox{\linewidth}{!}{
\begin{tabular}{l|ccc|c|ccc}
\toprule
\multirow{2}{*}{Condition Type} & \multicolumn{3}{c|}{Visual Quality} & Geo. Consist. & \multicolumn{3}{c}{Camera Accuracy} \\
 & PSNR$\uparrow$ & SSIM$\uparrow$ & LPIPS$\downarrow$ & MEt3R$\downarrow$ & mRotErr ($^\circ$)$\downarrow$ & mTransErr$\downarrow$ & mCamMC$\downarrow$ \\
\midrule
(i) Plücker Ray Camera Encoding              & 11.085 & 0.251 & {0.672} & 0.630 & 29.236 & {0.755} & {1.070} \\
(ii) (i) + Correspondence Loss                           & {11.186} & \textbf{0.282} & 0.666 & 0.435 & {13.585} & 0.725 & 0.837 \\
(iii) (i) + Tracking Module                  & 11.296 & {0.272} & {0.651} & {0.467} & 12.845 & 0.592 & 0.693 \\
\midrule
(iv) (i) + Tracking Module + Correspondence Loss         & \textbf{11.389} & 0.267 & \textbf{0.645} & \textbf{0.416} & \textbf{11.694} & \textbf{0.579} & \textbf{0.669} \\
\bottomrule
\end{tabular}
}
\vspace{-10pt}
\end{table}



\section{Conclusion}
We introduce \ours, which strengthens both motion and geometric correspondence in a video diffusion model for dynamic novel-view generation. Our key insight is that correspondence-specialized attention layers naturally emerge in novel-view diffusion models, jointly encoding intra-video temporal and inter-video cross-view correspondences — a property largely overlooked in prior works. We exploit this property with two complementary objectives that share the DiT's attention features: a multi-view point tracking objective that makes the generated view follow the reference motion, and a multi-view correspondence loss that enforces geometric consistency across and within views. Without any explicit 3D reconstruction at inference, our method achieves state-of-the-art geometric consistency over both reconstruction-based and reconstruction-free baselines.

\clearpage


\medskip

{\small
\bibliographystyle{plain}
\bibliography{references}
}

\appendix

\clearpage

\section{Correspondence in 3D Attention Map}

\subsection{Dataset for Analysis and Pseudo Ground-Truth Generation}
\label{sec:analysis_eval_details}
\paragraph{MultiCamVideo Dataset.}
We conduct our analysis (Sec.~\ref{sec:analysis}) on the MultiCamVideo dataset~\cite{bai2025recammaster}, a multi-camera synchronized video dataset rendered via Unreal Engine 5 that features large camera movements. The dataset provides time-synchronized multi-view recordings of dynamic scenes together with diverse camera trajectories. Since each scene is captured by multiple cameras at the same timestamps, the dataset naturally offers paired reference--target videos suitable for evaluating both temporal and multi-view correspondences. From this dataset, we sample 40 scenes, and for each scene we randomly select two camera views where one serves as the reference view and the other as the target view for novel-view generation.

\begin{figure}[t]
    \centering
    \includegraphics[width=\linewidth]{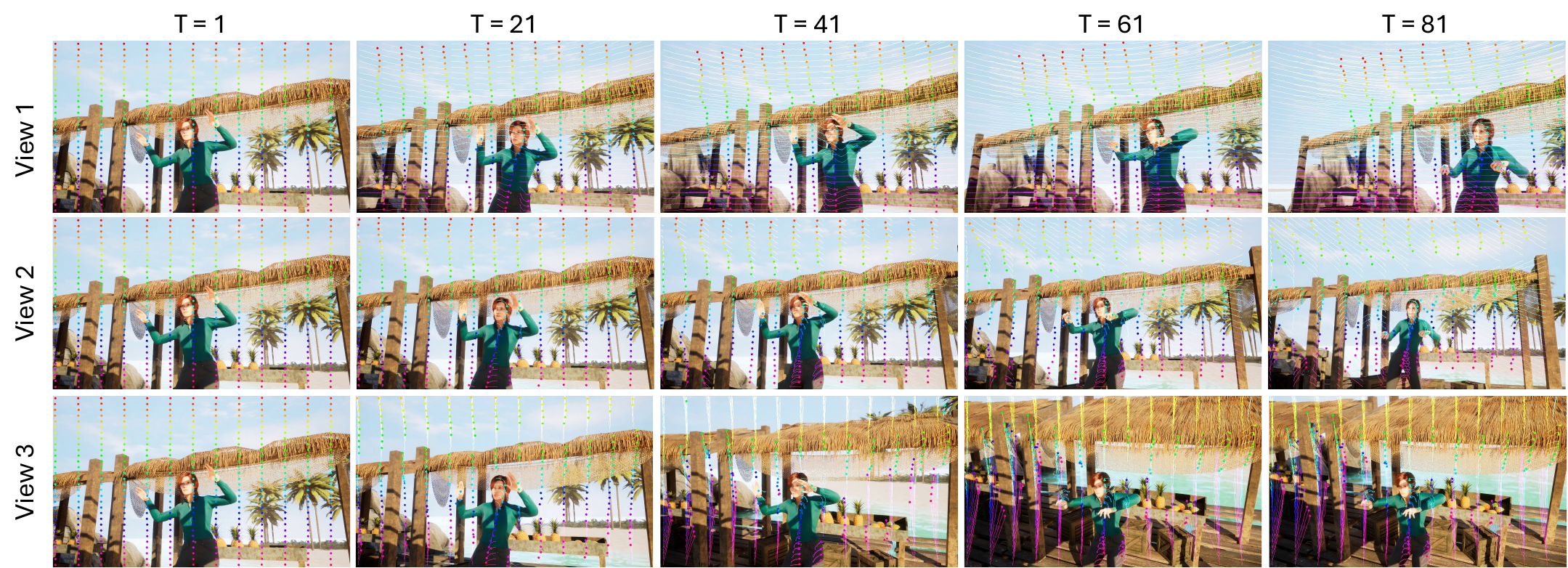}
    \caption{\textbf{Pseudo Ground-Truth Multi-View Tracks on the MultiCamVideo Dataset~\cite{bai2025recammaster}.}
    We run MV-TAP~\cite{koo2026mv} on the synchronized multi-view videos of the MultiCamVideo dataset to extract pseudo ground-truth multi-view point tracks. The tracked points are propagated consistently across both frames and views, providing intra-video temporal and inter-video cross-view correspondences that serve as supervision for training.}
    \label{supple_fig:recam_dataset}
\end{figure}

\paragraph{Pseudo Ground-Truth Multi-View Point Tracking via MV-TAP.}
To obtain dense point correspondences that span both the temporal axis within each video and the cross-view axis between the reference and target, we leverage MV-TAP~\cite{koo2026mv}, a recent multi-view point tracker that jointly reasons over synchronized views via cross-view attention. Concretely, we set query points on the first frame in a regular grid manner, and run MV-TAP jointly over the 10 synchronized view videos to estimate their correspondences across both time and viewpoints. Running MV-TAP on the reference--target pairs from the MultiCamVideo dataset yields pseudo multi-view point tracks $\mathcal{T} = \{ p^{v, \text{GT}}_{i} \}$ and visibility states $\mathcal{O} = \{ o^{v,\text{GT}}_{i} \}$, where $p^{v}_{i} \in \mathbb{R}^{N \times 2}$ denotes the 2D locations of $N$ tracked points in view $v$ at the $i$-th frame, and $o^{v}_{i} \in \{0, 1\}^{N}$ indicates whether each point is visible. These tracks serve as our pseudo ground-truth, as shown in Fig.~\ref{supple_fig:recam_dataset}.

\subsection{Details of Attention-based Correspondence Evaluation}
\label{sec:correspondence_details}

\paragraph{Forward Match via Mapping Operator.}
For a query point $p^{v_{1}}_{i}$ in the $i$-th latent frame of view $v_{1}$, its forward match in the $j$-th latent frame of view $v_{2}$ is obtained via $\mathrm{argmax}$ over the attention weight matrix $\mathcal{C}^{v_{1},v_{2}}_{i,j}$ in Eq.~\ref{eq:matching_cost}, which selects the spatial location with the highest attention score within the spatial domain $\Omega$ of the $j$-th latent. We define a mapping operator $\mathcal{F}$ that returns this forward match under a given attention weight matrix:
\begin{equation}
    \label{eq:mapping}
    \hat{p}^{v_2}_{j} = \mathcal{F}\bigl(\mathcal{C}^{v_{1},v_{2}}_{i,j},\, p^{v_1}_{i}\bigr) = \underset{p \in \Omega}{\mathrm{argmax}}\;\mathcal{C}^{v_{1},v_{2}}_{i,j}(p^{v_{1}}_{i},\, p), \quad v_{1},v_{2} \in \{ref, tgt\}.
\end{equation}
With this mapping, the forward match from view $v_1$ to view $v_2$ is $\hat{p}^{v_{2}}_{j} = \mathcal{F}(\mathcal{C}^{v_{1},v_{2}}_{i,j}, p^{v_{1}}_{i})$, and the backward match from $\hat{p}^{v_{2}}_{j}$ back to view $v_{1}$ is obtained by applying $\mathcal{F}$ once more with the reverse attention weight matrix $\mathcal{C}^{v_{2},v_{1}}_{j,i}$.

\paragraph{Reliable Correspondence via Cycle Consistency.}
To identify reliable correspondences in the attention map, we adopt a cycle-consistency check: a query and its forward match form a valid correspondence only when matching backward returns the original query. Formally, $p^{v_1}_{i}$ is regarded as reliable if
\begin{equation}
    \label{eq:reliable}
    \mathrm{Reliable}\bigl(p^{v_{1}}_{i};\, \mathcal{C}^{v_{1},v_{2}}_{i,j}\bigr) = \mathbbm{1}\!\left[\, \bigl\lVert \mathcal{F}\bigl(\mathcal{C}^{v_{2},v_1}_{j,i},\, \mathcal{F}(\mathcal{C}^{v_{1},v_{2}}_{i,j},\, p^{v_{1}}_{i})\bigr) - p^{v_{1}}_{i} \bigr\rVert_2 \le \delta \,\right]
\end{equation}
where $\delta$ denotes the cycle-consistency threshold on the pixel grid and we set $\delta = 16$.
This enforces a mutual best-match relation: both directions of attention should agree on the same point pair. Since the attention is computed in the latent space, the pseudo ground-truth tracks $\mathcal{T}$ are accordingly rescaled to the latent resolution.


\begin{figure}[t]
    \centering
    \includegraphics[width=0.9\linewidth]{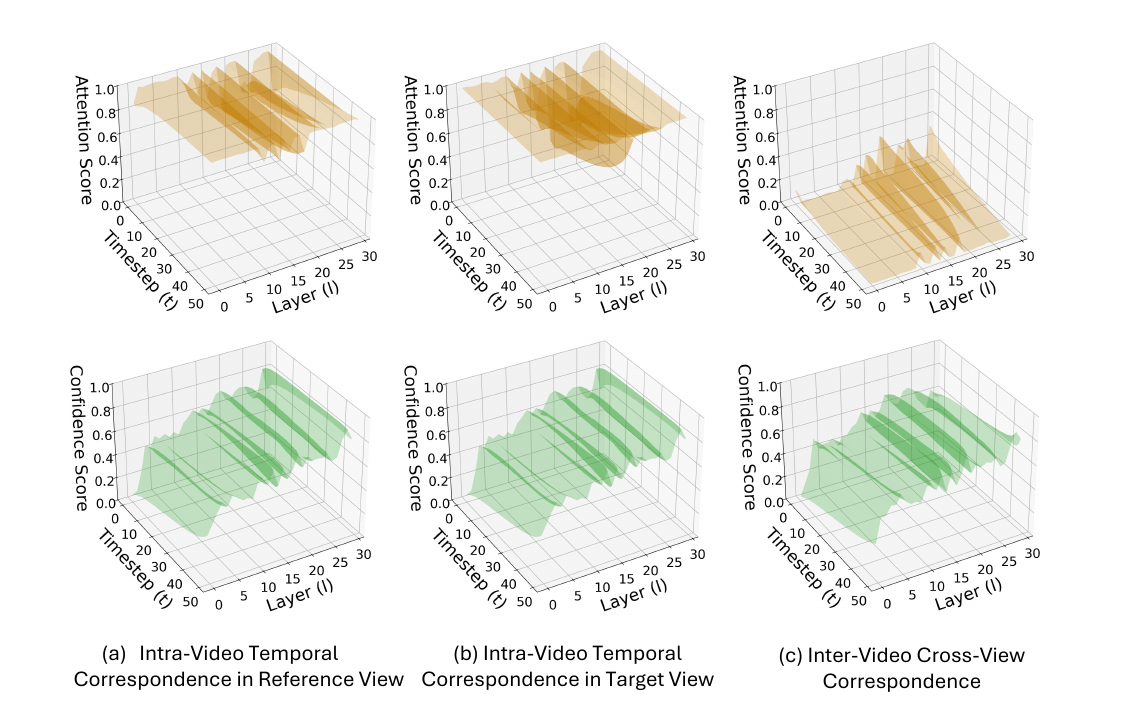}
    \caption{\textbf{Attention Score and Confidence Score in ReCamMaster~\cite{bai2025recammaster}.}
    We visualize the attention score (top) and confidence score (bottom) of query points across diffusion layers $l$ and timesteps $t$ for the three correspondence types: (a) intra-video temporal in the reference view, (b) intra-video temporal in the target view, and (c) inter-video cross-view.}
    \label{supple_fig:attention_score_and_condifence_recam}
\end{figure}

\paragraph{Matching Accuracy via PCK.}
We measure matching accuracy using the Percentage of Correct Keypoints (PCK). Given the pseudo ground-truth tracks $\mathcal{T}$ rescaled to the latent grid, a query point $p^{v_{1}}_{i}$ is counted as a positive when both conditions hold: (i) it is reliable under Eq.~\ref{eq:reliable}, and (ii) its forward match $\hat{p}^{v_2}_{j} = \mathcal{F}(\mathcal{C}^{v_1,v_2}_{i,j}, p^{v_1}_{i})$ lies  within $\delta$ pixels of the co-visible ground-truth match $p^{v_2,\text{GT}}_{j}$ on the latent grid. Formally,
\begin{equation}
    \label{eq:pck}
    \mathrm{PCK} = \frac{1}{|\mathcal{Q}|} \sum_{p^{v_1}_{i} \in \mathcal{Q}} \mathbbm{1}\!\left[\, \mathrm{Reliable}\bigl(p^{v_1}_{i};\, \mathcal{C}^{v_1,v_2}_{i,j}\bigr) = 1 \;\wedge\; \bigl\lVert \hat{p}^{v_2}_{j} - p^{v_2,\text{GT}}_{j} \bigr\rVert_2 \le \delta \,\right],
\end{equation}
where $\mathcal{Q}$ is the set of all query points whose ground-truth match is co-visible across the relevant frames. Matches are obtained on the latent grid, while distances are measured in the original pixel space with $\delta$ set to $16$ pixels. PCK is computed separately for each of the three correspondence types and averaged across latent frames and scenes.

\paragraph{Attention Score.}
A single query point attends to tokens from both the reference and target views through the attention mechanism, allowing us to examine how its attention weights concentrate across layers and denoising timesteps. The resulting attention map encodes three types of correspondence: (a) intra-video temporal correspondence within the reference view, (b) intra-video temporal correspondence within the target view, and (c) inter-video cross-view correspondence between the two views. For each type, we visualize the attention weights of individual query points. As shown in the top row of Fig.~\ref{supple_fig:attention_score_and_condifence_recam}, the temporal correspondences within each view, shown in (a) and (b), are emphasized in the early and final diffusion layers, whereas in the intermediate layers the attention shifts toward the cross-view correspondence in (c).

\paragraph{Confidence Score.}
The confidence score captures \emph{how sharply} the correspondence is localized within the 3D attention map. For each query point, we re-normalize the attention over only the keys of the ground-truth frame and take the maximum logit value of the attention map, which approaches one when the distribution collapses onto a single key and stays low when it remains diffuse. As shown in the bottom row of Fig.~\ref{supple_fig:attention_score_and_condifence_recam}, the temporal correspondences within each view, shown in (a) and (b), tend to become sharply localized in the later diffusion layers, whereas the cross-view correspondence in (c) exhibits higher confidence in the intermediate layers.

\subsection{Generalization to Another Backbone}
\label{sec:redirector_analysis}
To verify that the correspondence-specialized layer structure is not an artifact of a single model, we repeat the same analysis on Redirector~\cite{park2025redirector}, a camera-controlled video diffusion backbone. We compute matching accuracy, attention score, and confidence score, and average the three into a harmonic mean, using identical settings to those used for ReCamMaster~\cite{bai2025recammaster}, and report them over various diffusion layers $l$ and denoising timesteps $t$ for the three correspondence types.

\begin{figure}[p]
    \centering
    \includegraphics[width=\linewidth]{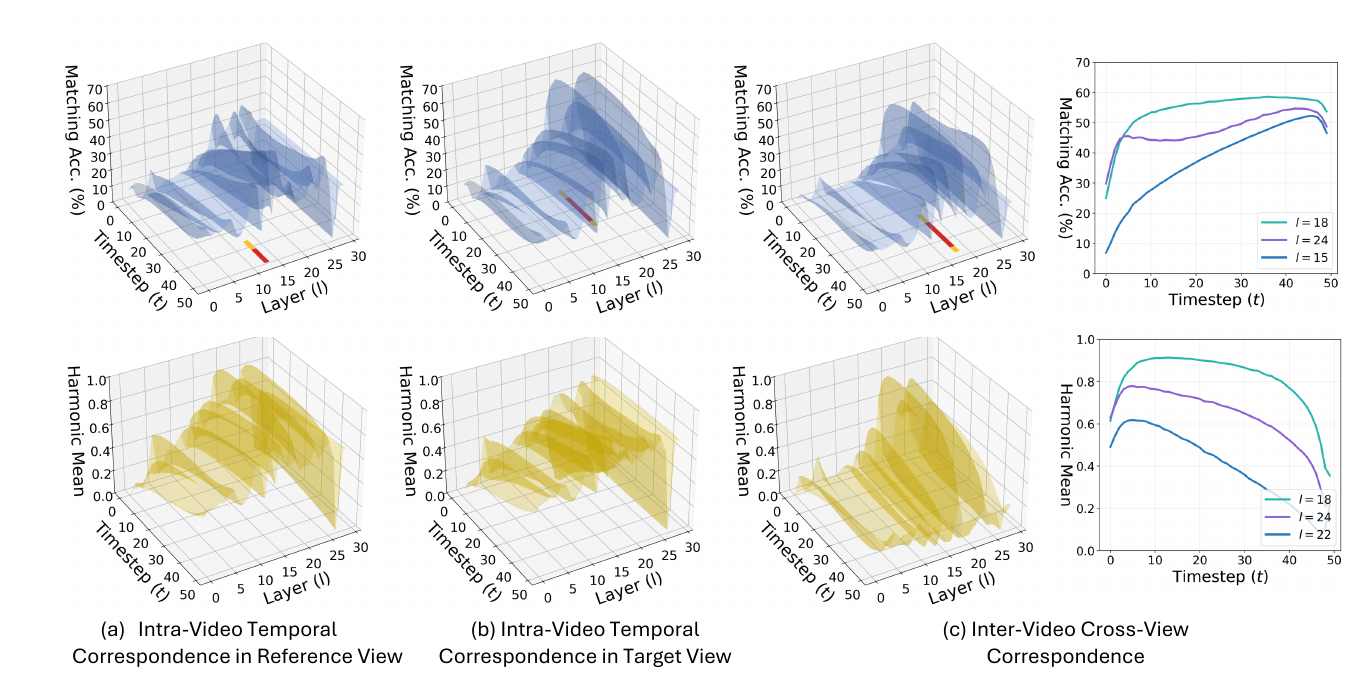}
    \caption{\textbf{Accuracy and Harmonic Mean in Redirector~\cite{park2025redirector}.} We visualize matching accuracy (top) and the harmonic mean of accuracy, confidence, and attention score (bottom) over various diffusion layers and denoising timesteps, for (a) intra-video temporal correspondence in the reference view, (b) intra-video temporal correspondence in the target view, and (c) inter-video cross-view correspondence. The right column shows selected layers from (c) plotted against timestep $t$. Consistent with our analysis on ReCamMaster, cross-view matching and temporal matching within the reference and target views emerge at specific middle layers and become stronger as denoising progresses, indicating that the correspondence-specialized layer structure is not specific to a single backbone.}
    \label{supple_fig:attention_accuracy_harmonic_mean_redirector}

    \vspace{1em}

    \includegraphics[width=0.9\linewidth]{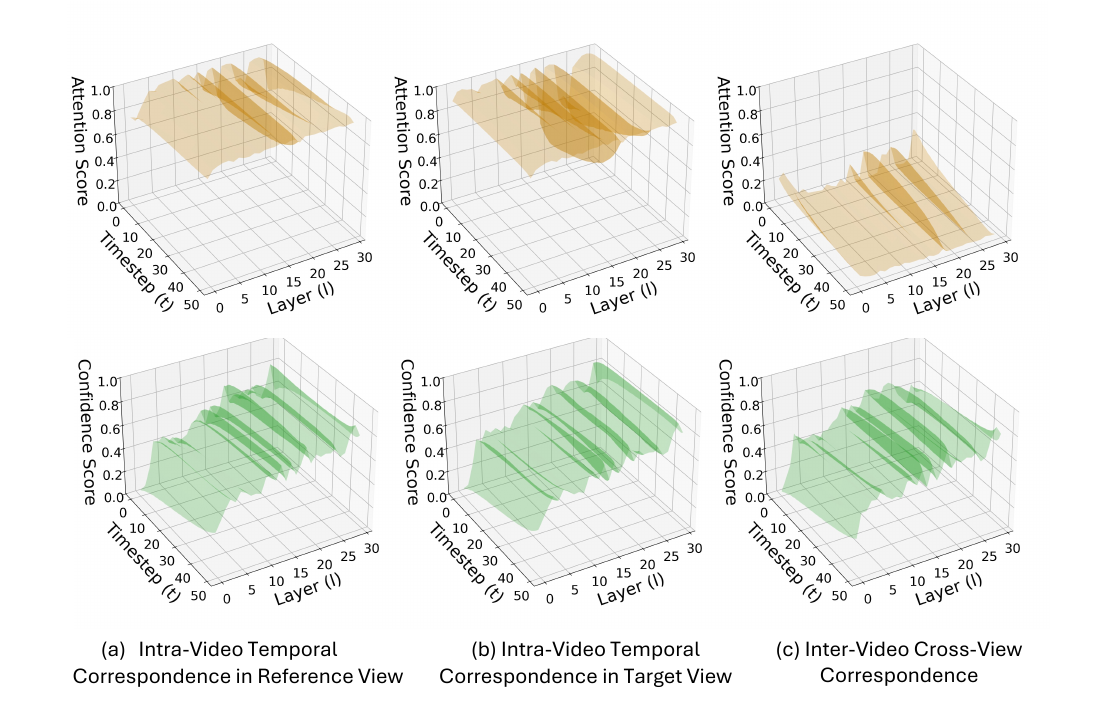}
    \caption{\textbf{Attention Score and Confidence Score in Redirector~\cite{park2025redirector}.} We visualize the attention score (top) and confidence score (bottom) of query points across diffusion layers $l$ and timesteps $t$, for (a) intra-video temporal correspondence in the reference view, (b) intra-video temporal correspondence in the target view, and (c) inter-video cross-view correspondence. As on ReCamMaster, the temporal correspondences (a, b) are emphasized in the early and final layers while the cross-view correspondence (c) concentrates in the intermediate layers, confirming that the same layer-wise behavior holds across backbones.}
    \label{supple_fig:attention_score_confidence_redirector}
\end{figure}

\paragraph{Accuracy and Harmonic Mean.}
As shown in Fig.~\ref{supple_fig:attention_accuracy_harmonic_mean_redirector}, the matching accuracy (top) and the harmonic mean (bottom) on Redirector follow the same layer–timestep trend observed on ReCamMaster: intra-video temporal correspondences in both the reference and target views, shown in (a) and (b), peak in a similar intermediate layer range, whereas the inter-video cross-view correspondence in (c) is more sharply localized around a specific middle 18th layer. The selected-layer curves in the right column further show that matching becomes accurate as denoising progresses. This consistency indicates that the emergence of a correspondence-specialized layer is a property shared across camera-controlled video diffusion backbones rather than one tied to a particular model.

\paragraph{Attention Score.}
As shown in the top row of Fig.~\ref{supple_fig:attention_score_confidence_redirector}, Redirector exhibits a similar attention-routing pattern to ReCamMaster. Intra-video temporal correspondences in the reference and target views, shown in (a) and (b), receive stronger attention in the early and later diffusion layers, while inter-video cross-view correspondence in (c) becomes more prominent in the intermediate layers. This consistent trend suggests that camera-controlled video diffusion backbones tend to allocate intermediate layers to cross-view information exchange.

\paragraph{Confidence Score.}
The confidence maps in the bottom row of Fig.~\ref{supple_fig:attention_score_confidence_redirector} further support this observation. Temporal correspondences become more sharply localized toward the later layers, whereas cross-view correspondences show higher confidence around the intermediate layer range. Together with the attention-score analysis, these results indicate that Redirector also develops a correspondence-specialized layer structure, supporting the generality of our choice to supervise attention features from this layer range.



\clearpage
\clearpage

\section{Model Architecture}
\subsection{Plücker Ray Camera Encoding}
\label{sec:camera_details}

\paragraph{Pl\"ucker Ray Construction.}
We provide the formal construction of the Pl\"ucker ray map. For each pixel, the corresponding ray $\mathbf{r} \in \mathbb{R}^{6}$ is defined as
\begin{equation}
    \label{eq:plucker}
    \mathbf{r} = \begin{bmatrix} \mathbf{d} \\ \mathbf{m} \end{bmatrix}, \quad \text{where } \mathbf{m} = \mathbf{o} \times \mathbf{d},
\end{equation}
with the ray direction $\mathbf{d} \in \mathbb{R}^{3}$ and origin $\mathbf{o} \in \mathbb{R}^{3}$ computed as
\begin{equation}
    \mathbf{d} = \mathbf{R}^{\top}\mathbf{K}^{-1}\mathbf{x}, \quad \mathbf{o} = -\mathbf{R}^{\top}\mathbf{t},
\end{equation}
where $\mathbf{x} = (u, v, 1)^{\top}$ is the homogeneous pixel coordinate, $\mathbf{R}$ and $\mathbf{t}$ are the camera rotation and translation, and $\mathbf{K}$ is the intrinsic matrix. The direction $\mathbf{d}$ is normalized to unit length for scale invariance. Aggregating the rays over all pixels and frames yields the dense Pl\"ucker maps $\mathbf{r}_{\text{ref}}, \mathbf{r}_{\text{tgt}} \in \mathbb{R}^{6 \times F \times H \times W}$.

\paragraph{Incorporating Pl\"ucker Maps into DiT Layers.}
The Pl\"ucker maps are injected into each DiT layer as
\begin{equation}
    h' = h + \mathrm{Proj}\bigl(\mathrm{CamEnc}([\mathbf{r}_{\text{ref}}, \mathbf{r}_{\text{tgt}}])\bigr),
\end{equation}
where $h$ is the input token to the 3D attention at each layer, $\mathrm{CamEnc}$ is a lightweight convolutional encoder that maps the 6-channel Pl\"ucker rays to the token dimension, and $\mathrm{Proj}$ is a linear projection. We follow the same injection location as ReCamMaster~\cite{bai2025recammaster}.

\subsection{Multi-Scale Local 4D Correlation}
\label{sec:hierarchical_local_4d_correlation}
We construct a multi-scale local 4D correlation with varying receptive fields to capture multi-scale query--key relationships from the 3D attention. To build a feature pyramid with $S$ scales, we interpolate the DiT query and key features $Q^{v}_{i}, K^{v}_{i}$ at each transformer layer to resolution $\tfrac{H}{r \times 2^{s-1}} \times \tfrac{W}{r \times 2^{s-1}}$, where $r=4$ denotes the model stride and $s \in \{1,2,3,4\}$ denotes the scale factor:
\begin{align}
    {}^{s}Q^{v}_{i} &= \mathrm{Interpolate}_{s}(Q^{v}_{i}), \\
    {}^{s}K^{v}_{i} &= \mathrm{Interpolate}_{s}(K^{v}_{i}), \quad v \in \{\text{ref}, \text{tgt}\}.
\end{align}
At each scale $s$, we extract local features around points of interest via bilinear sampling. Local query features ${}^{s}q^{v}_{i}$ are sampled within a $\Delta$-sized neighborhood centered at the query point $p^{v}_{i} = (x^{v}_{i}, y^{v}_{i})$ in frame $i$ of view $v$:
\begin{equation}
    {}^{s}q^{v}_{i} = {}^{s}Q^{v}_{i}\!\left(\frac{x^{v}_{i}}{2^{s-1}} + \delta_{x},\; \frac{y^{v}_{i}}{2^{s-1}} + \delta_{y} : \|\delta\|_{\infty} \leq \Delta \right),
\end{equation}
and local key features ${}^{s}k^{v}_{j}$ are sampled around the estimated point $\hat{p}^{v}_{j} = (\hat{x}^{v}_{j}, \hat{y}^{v}_{j})$ across all frame indices $j$ in view $v$:
\begin{equation}
    {}^{s}k^{v}_{j} = {}^{s}K^{v}_{j}\!\left(\frac{\hat{x}^{v}_{j}}{2^{s-1}} + \delta_{x},\; \frac{\hat{y}^{v}_{j}}{2^{s-1}} + \delta_{y} : \|\delta\|_{\infty} \leq \Delta \right),
\end{equation}
where $\delta \in \mathbb{Z}^{2}$ and ${}^{s}q^{v}_{i}, {}^{s}k^{v}_{j} \in \mathbb{R}^{d \times (2\Delta+1)^2}$, with $v \in \{\text{ref}, \text{tgt}\}$.
We then construct a local 4D correlation volume ${}^{s}\operatorname{Corr}^{v}_{i,j}$ between local features of frame $i$ and frame $j$ in view $v$ via the softmax operation:
\begin{equation}
    {}^{s}\operatorname{Corr}^{v}_{i,j} = \mathrm{Softmax}\!\left(\frac{{}^{s}q^{v}_{i} \bigl({}^{s}k^{v}_{j}\bigr)^{\top}}{\sqrt{d_{\text{head}}}}\right) \in \mathbb{R}^{(2\Delta+1)^4}, \quad v \in \{\text{ref}, \text{tgt}\}.
\end{equation}

We concatenate the local 4D correlation volumes from all $S$ scales along the channel dimension to obtain the multi-scale correlation descriptor $\operatorname{Corr}^{v}_{i,j} \in \mathbb{R}^{4(2\Delta+1)^4}$.

\begin{equation}
    \operatorname{Corr}^{v}_{i,j} = \operatorname{Concat}\!\Big(
    {}^{1}\operatorname{Corr}^{v}_{i,j},\,
    {}^{2}\operatorname{Corr}^{v}_{i,j},\,
    {}^{3}\operatorname{Corr}^{v}_{i,j},\,
    {}^{4}\operatorname{Corr}^{v}_{i,j}
    \Big) \in \mathbb{R}^{4(2\Delta+1)^4}.
\end{equation}

\subsection{Layer Ablation}
\label{sec:layer_ablation}
Our analysis in Sec.~\ref{sec:analysis} identifies the 18th DiT layer as the correspondence-specialized layer whose 3D attention most reliably encodes inter-video cross-view correspondence. This observation translates into the best downstream generation, i.e., that the layer carrying the strongest correspondence signal is also the most effective source of features for our multi-view tracking head. We ablate the matching layer~$l$ --- the DiT layer whose query and key features are routed into the tracking head --- across four layers spanning the network, and report generation quality, geometric consistency, and camera accuracy on the MultiCamVideo~\cite{bai2025recammaster} evaluation set in Table~\ref{tab:layer_ablation}.
Layer~$18$ attains the best generation quality on PSNR, SSIM, and LPIPS, the lowest MEt3R, and the most accurate camera control on all three pose metrics, consistent with our analysis where cross-view correspondence peaks at this layer. 

\begin{table}[t]
    \centering
    \caption{\textbf{Layer ablation on the MultiCamVideo~\cite{bai2025recammaster} dataset.} We vary the matching layer~$l$ that feeds the multi-view tracking head and the best value per metric is shown in \textbf{bold}.}
    \label{tab:layer_ablation}
    \setlength{\tabcolsep}{3pt}
    \footnotesize
    \resizebox{\linewidth}{!}{
    \begin{tabular}{l|ccc|c|ccc}
    \toprule
    \multirow{2}{*}{Matching Layer} & \multicolumn{3}{c|}{Visual Quality} & Geo. Consist. & \multicolumn{3}{c}{Camera Accuracy} \\
    & PSNR$\uparrow$ & SSIM$\uparrow$ & LPIPS$\downarrow$ & MEt3R$\downarrow$ & mRotErr ($^\circ$)$\downarrow$ & mTransErr$\downarrow$ & mCamMC$\downarrow$ \\
    \midrule
    \quad $l=15$            & 12.78 & 0.505 & 0.573 & 0.451 & 15.58 & 67.94 & 72.98 \\
    \quad \textbf{$l=18$} & \textbf{14.08} & \textbf{0.559} & \textbf{0.496} & \textbf{0.394} & \textbf{4.55} & \textbf{20.31} & \textbf{22.35} \\
    \quad $l=22$            & 12.38 & 0.484 & 0.558 & 0.465 &  9.61 & 46.05 & 50.85 \\
    \quad $l=24$            &  9.99 & 0.422 & 0.708 & 0.469 & 25.71 & 74.61 & 85.57 \\
    
    \bottomrule
    \end{tabular}
    }
\end{table}

\subsection{Training Data}
\label{sec:training_data}

We train on a hybrid dataset combining three complementary sources to maximize diversity in scene types, camera motions, and visual domains:

\paragraph{Kubric.}
The Kubric dataset provides synthetic multi-object scenes rendered with Blender, precise camera intrinsics, extrinsics, and dense point tracks with ground-truth visibility labels.
Videos are center-cropped from $832{\times}832$ to $480{\times}832$ at 81 frames.

\paragraph{MultiCamVideo.}
The MultiCamVideo dataset provides cinematic multi-camera sequences rendered via Unreal Engine~5, featuring 10 synchronized camera viewpoints per scene with known calibration. Dense point tracks are computed on a $30{\times}52$ spatial grid (1{,}560 points per frame).


\paragraph{Reverse MultiCamVideo Augmentation.}
To improve robustness, we augment training by reversing the temporal order of the video sequences, which increases the diversity of both object motion and camera trajectories seen during training.

\subsection{Training Objective Details}
\label{sec:supp_loss}


\paragraph{Diffusion Loss.} 
We adopt the rectified flow formulation,
\begin{equation}
    \mathcal{L}_{\text{diff}} = w(t)\,\bigl\| \mathbf{v}_\theta\!\left([\, z_{\text{ref}},\, z_{\text{tgt}} \,], t, c, \mathrm{cam}_{\text{ref}},\mathrm{cam}_{\text{tgt}}\right) - (\epsilon - x_0) \bigr\|^2_{\text{tgt}},
\end{equation}
where $\mathbf{v}_\theta(\cdot)$ is the predicted velocity field conditioned on the concatenated reference and target latents $[z_{\text{ref}}, z_{\text{tgt}}]$, the denoising timestep $t \sim \mathcal{U}(0, 1)$, the conditioning signal $c$, and the target camera $\mathrm{cam}_{\text{tgt}}$. The training target is the rectified flow velocity $\epsilon - x_0$, where $\epsilon \sim \mathcal{N}(0, I)$ is the sampled noise and $x_0$ denotes the clean target latent, with the noisy input constructed as $x_t = (1-t)\,x_0 + t\,\epsilon$. The weighting $w(t)$ is a timestep-dependent factor that balances the contribution of different noise levels during training. The squared error is evaluated only on the target-view tokens, as indicated by the $\|\cdot\|^2_{\text{tgt}}$ notation; reference-view tokens serve purely as conditioning context and do not contribute to the diffusion loss.

\paragraph{Tracking Loss.} The multi-view point tracking head~\cite{koo2026mv} is supervised with ground-truth point trajectories using
\begin{equation}
    \mathcal{L}_{\text{track}} = \lambda_{\text{seq}}\,\mathcal{L}_{\text{seq}} + \lambda_{\text{conf}}\,\mathcal{L}_{\text{conf}} + \lambda_{\text{vis}}\,\mathcal{L}_{\text{vis}},
\end{equation}
with $\lambda_{\text{seq}}=0.05$ and $\lambda_{\text{conf}}=\lambda_{\text{vis}}=1.0$. Given the ground-truth tracks $\mathcal{T} = \{p^{v,\text{GT}}_{n,i}\}$ and visibility $\mathcal{O} = \{o^{v,\text{GT}}_{n,i} \in \{0, 1\}\}$ of the $N$ query points indexed by $n$ over the $i$ latent frames of both views $v \in \{\text{ref}, \text{tgt}\}$, together with the head's predictions $\{\hat{p}^{v}_{n,i}, \hat{c}^{v}_{n,i}, \hat{o}^{v}_{n,i}\}$, the three terms are defined as follows. The sequence loss $\mathcal{L}_{\text{seq}}$ is a visibility-weighted Huber loss~\cite{huber1992robust} on the predicted 2D point coordinates,
\begin{equation}
    \mathcal{L}_{\text{seq}} = \frac{1}{\sum_{v,n,i} o^{v,\text{GT}}_{n,i}}\sum_{v,n,i} o^{v,\text{GT}}_{n,i}\,\rho\!\left(\hat{p}^{v}_{n,i} - p^{v,\text{GT}}_{n,i}\right),
\end{equation}
where $\rho(\cdot)$ is the Huber function; this restricts coordinate regression to frames where the ground-truth point is visible and prevents occluded targets from injecting noisy gradients. The confidence loss $\mathcal{L}_{\text{conf}}$ is a probabilistic loss that calibrates the predicted per-point confidence $\hat{c}^{v}_{n,i}$ against the realized regression error: confidences are supervised to be high when the predicted location lies within a small radius of the ground-truth and low otherwise, encouraging the head to produce reliability estimates that downstream consumers can use to filter unreliable tracks.

\paragraph{Multi-View Correspondence Loss.}
While the tracking head is supervised only through the sampled local correlation volumes in each view, the attention weight matrix $\mathcal{C}^{v_1,v_2}_{i,j}$ in Eq.~\ref{eq:matching_cost} itself encodes three types of correspondence within the attention map: \emph{intra-video temporal correspondences} within the reference view and within the target view, and \emph{inter-video cross-view correspondence} between the two views. These correspondences act as a bridge that exchanges information both across views and across time, and visually failed regions in the generated output tend to coincide with correspondences that are misaligned in the attention map. We therefore supervise it directly with a cross-entropy objective derived from the multi-view point tracks $\mathcal{T}$, so that each query point attends to the location where it physically reappears in the other frames and the other view. The key property we exploit is that the multi-view tracks specify the correct token \emph{exactly}: for a query point $p^{v_1}_{i}$ in the $i$-th latent frame of view $v_1$, its match in any other latent frame $j$ of view $v_2$ is either a single ground-truth location $p^{v_2,\text{GT}}_{j}$ or occluded, as indicated by the visibility flag $o^{v_2,\text{GT}}_{j} \in \{0, 1\}$.

We randomly sample a set of query points $\mathcal{Q}$, and for each query we supervise its correspondence against every co-visible target token across all latent frames $j$ of both the reference and target views $v_2 \in \{\text{ref}, \text{tgt}\}$. Since each row $\mathcal{C}^{v_1,v_2}_{i,j}(p^{v_1}_{i}, \cdot)$ is already a per-frame softmax distribution over the keys of the $j$-th latent frame, we apply a visibility-weighted cross-entropy loss for each query point $p^{v_1}_{i}$:
\begin{equation}
    \label{eq:ce_loss}
    \mathcal{L}_{\text{corr}}
    =
    \sum_{v_2} \sum_{j}
    o^{v_2,\text{GT}}_{j}\,
    \text{CE}\left(
        \mathcal{C}^{v_1,v_2}_{i,j}\!\left(p^{v_1}_{i}, \cdot\right),
        p^{v_2,\text{GT}}_{j}
    \right).
\end{equation}
Here, $\text{CE}(\cdot,\cdot)$ denotes the cross-entropy loss with the ground-truth matching point $p^{v_2,\text{GT}}_{j}$ as the target label. By averaging over all co-visible target frames $j$ in both views $v_2 \in \{\text{ref}, \text{tgt}\}$, the loss jointly supervises intra-video temporal correspondences within each view and inter-video cross-view correspondences across the two views, directly shaping where each query token attends to enforce geometric consistency both across the reference and generated views and within the generated views themselves.




\section{More Generation Results}
\label{sec:more_gen_qual}
\begin{figure*}[ht!]
    \centering
    \includegraphics[width=\linewidth]{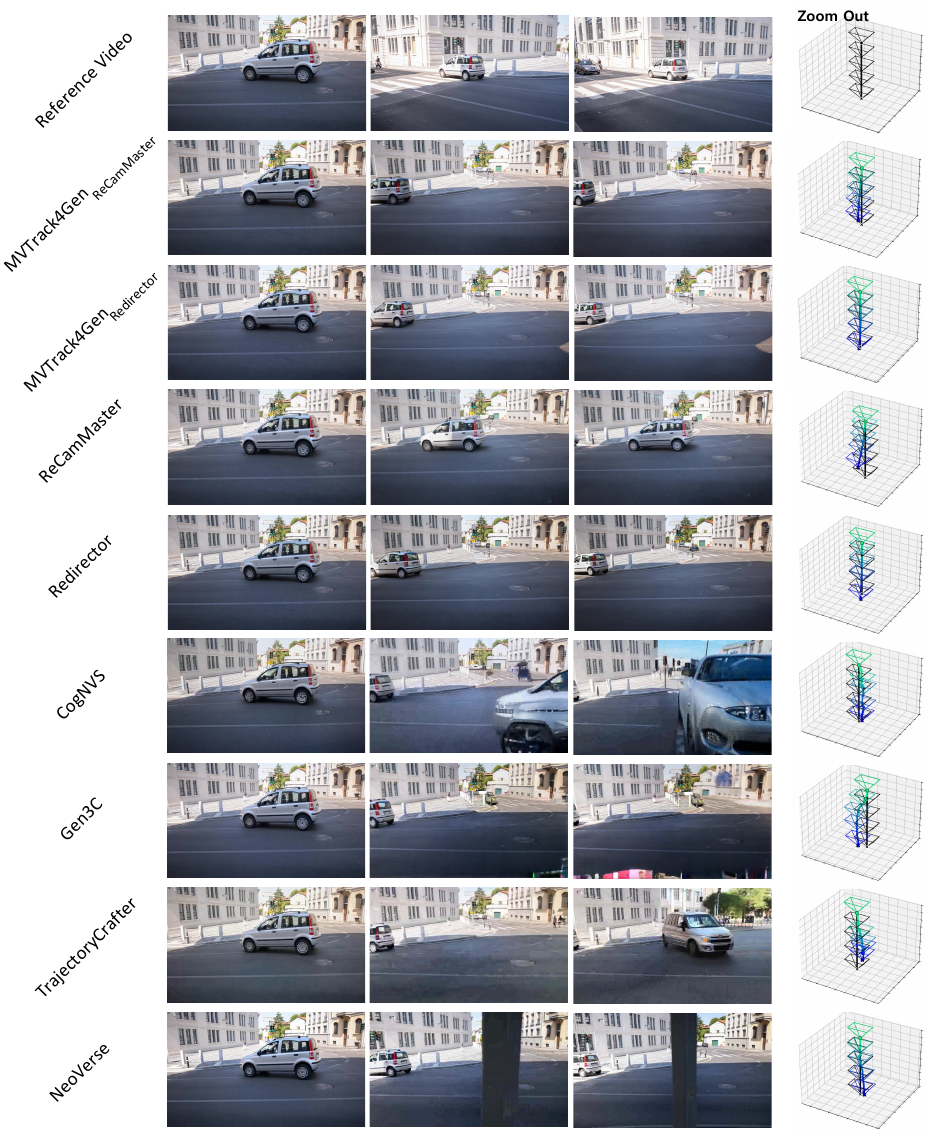}
    \caption{\textbf{Additional qualitative comparisons on the DAVIS dataset~\cite{perazzi2016benchmark}.}
    The top {Reference Video} row shows reference-view frames at three timesteps, each row below renders the same target viewpoint, and the rightmost \textit{Zoom Out} column visualizes the target camera trajectory in 3D. We compare our two models, \ours\textsubscript{ReCamMaster} and \ours\textsubscript{Redirector}, against  ReCamMaster~\cite{bai2025recammaster}, Redirector~\cite{park2025redirector} and CogNVS~\cite{chen2025reconstruct}, GEN3C~\cite{ren2025gen3c}, TrajectoryCrafter~\cite{yu2025trajectorycrafter}, NeoVerse~\cite{yang2026neoverse}. Baselines show geometric distortions on dynamic foreground objects and floating or blurry background artifacts under large viewpoint changes, while \ours\ produces sharp, geometrically consistent novel-views faithful to the reference video.
    }
    \label{fig:supple_qual_1}
\end{figure*}

\begin{figure*}[ht!]
    \centering
    \includegraphics[width=\linewidth]{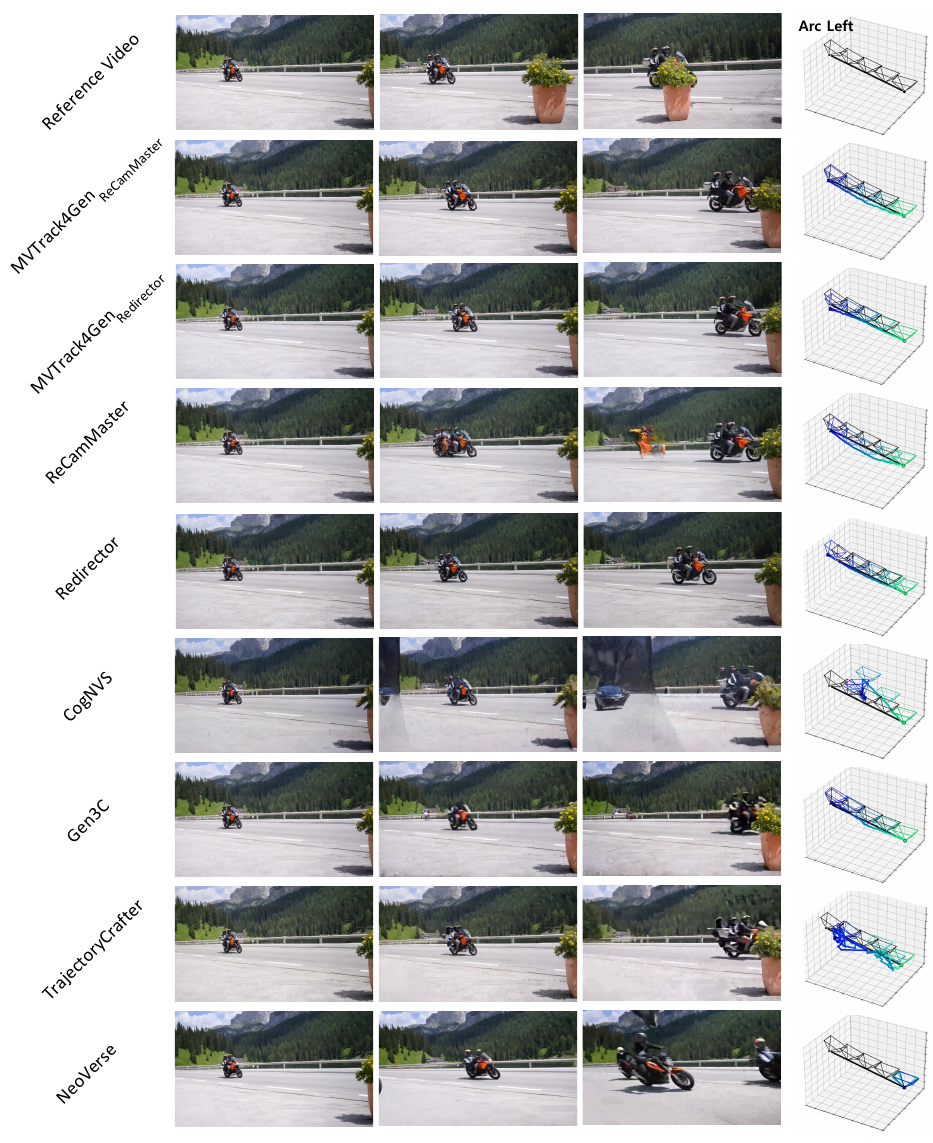}
    \caption{\textbf{Additional qualitative comparisons on the DAVIS dataset~\cite{perazzi2016benchmark}.}
    The top {Reference Video} row shows reference-view frames at three timesteps, each row below renders the same target viewpoint, and the rightmost \textit{Arc Left} column visualizes the target camera trajectory in 3D. We compare our two models, \ours\textsubscript{ReCamMaster} and \ours\textsubscript{Redirector}, against  ReCamMaster~\cite{bai2025recammaster}, Redirector~\cite{park2025redirector} and CogNVS~\cite{chen2025reconstruct}, GEN3C~\cite{ren2025gen3c}, TrajectoryCrafter~\cite{yu2025trajectorycrafter}, NeoVerse~\cite{yang2026neoverse}. Baselines show geometric distortions on dynamic foreground objects and floating or blurry background artifacts under large viewpoint changes, while \ours\ produces sharp, geometrically consistent novel-views faithful to the reference video.
    }
    \label{fig:supple_qual_3}
\end{figure*}

\begin{figure*}[ht!]
    \centering
    \includegraphics[width=\linewidth]{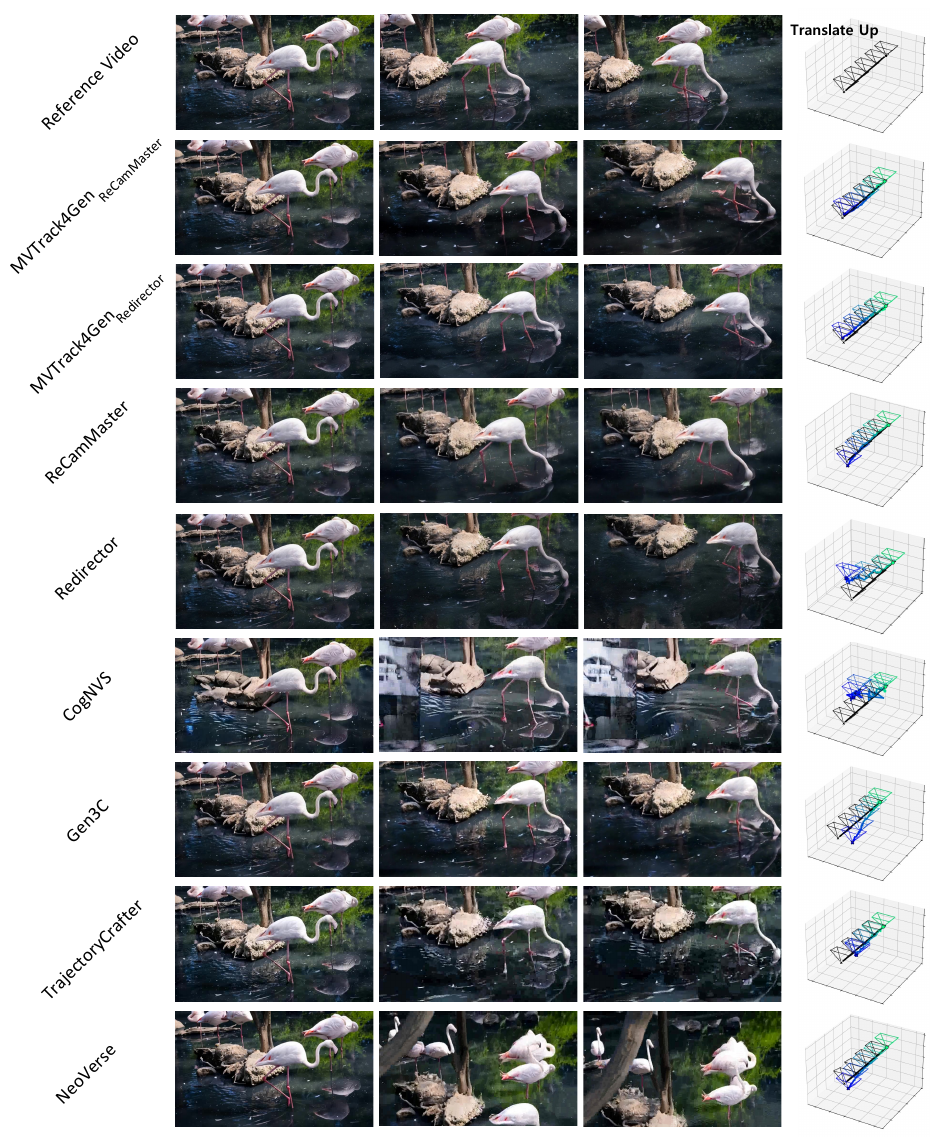}
    \caption{\textbf{Additional qualitative comparisons on the DAVIS dataset~\cite{perazzi2016benchmark}.}
    The top {Reference Video} row shows reference-view frames at three timesteps, each row below renders the same target viewpoint, and the rightmost \textit{Translate Up} column visualizes the target camera trajectory in 3D. We compare our two models, \ours\textsubscript{ReCamMaster} and \ours\textsubscript{Redirector}, against  ReCamMaster~\cite{bai2025recammaster}, Redirector~\cite{park2025redirector} and CogNVS~\cite{chen2025reconstruct}, GEN3C~\cite{ren2025gen3c}, TrajectoryCrafter~\cite{yu2025trajectorycrafter}, NeoVerse~\cite{yang2026neoverse}. Baselines show geometric distortions on dynamic foreground objects and floating or blurry background artifacts under large viewpoint changes, while \ours\ produces sharp, geometrically consistent novel-views faithful to the reference video.
    }
    \label{fig:supple_qual_4}
\end{figure*}

To complement the iPhone results in Fig.~\ref{fig:iphone_main_qual}, we provide additional
qualitative comparisons on the DAVIS dataset~\cite{perazzi2016benchmark}, which contains
in-the-wild monocular videos with diverse object and camera motion.
Figs.~\ref{fig:supple_qual_1},~\ref{fig:supple_qual_3}, and~\ref{fig:supple_qual_4}
show novel-view generations across representative scenes covering both dynamic foreground
subjects and wide outdoor environments, each under a different target camera trajectory.
\textit{Explicit 3D Lifting} baselines such as CogNVS, GEN3C, and TrajectoryCrafter preserve
the coarse scene layout but introduce noticeable artifacts in regions where depth estimation
becomes unreliable, including water surfaces, thin structures, and fast-moving subjects.
NeoVerse mitigates some of these artifacts but still produces residual distortions on
dynamic objects under large camera motion.
\textit{Camera Conditioning Only} baselines such as ReCamMaster and Redirector synthesize
plausible-looking frames but lack geometric grounding under viewpoint change, leading to
inaccurate parallax, drifting backgrounds, and non-rigid warping of foreground objects.
In contrast, our \ours\ generates novel-views that remain visually faithful to the reference
video while accurately following the prescribed target camera trajectory, preserving both
the structure of dynamic subjects and the geometry of the surrounding scene.
These qualitative observations are consistent with the quantitative trends reported in
Tab.~\ref{tab:davis_results}, further confirming that strengthening multi-view correspondence
learning into a video diffusion model is effective for in-the-wild novel-view synthesis,
even without any explicit 3D representation at inference.

\section{Attention Visualization Results}
\label{supple_sec:more_attention_vis}
\begin{figure*}[t]
    \centering
    \includegraphics[width=\linewidth]{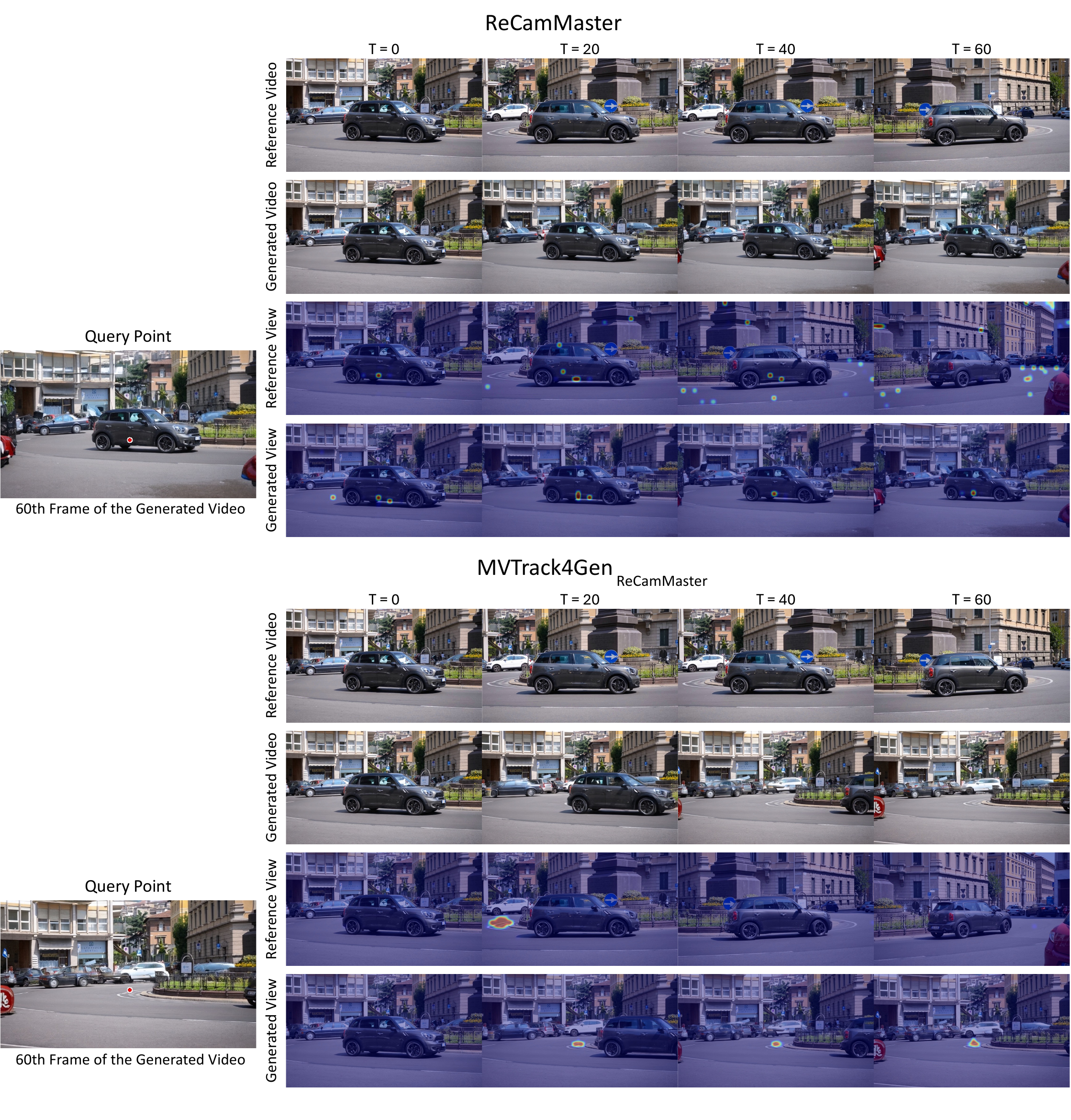}
    \caption{\textbf{Correlation between generation quality and attention-map alignment.}
    For both ReCamMaster~\cite{bai2025recammaster} (top) and our \ours\ (bottom), we show, from top to bottom, the {reference video}, the generated {novel-view video}, and the corresponding attention maps overlaid on the \emph{reference view} and the \emph{generated view} across frames.
    Given a query point placed on the 60th frame of the generated video (left), we visualize the attention distribution extracted from the 18th layer, identified as the correspondence-specialized layer in Sec.~\ref{sec:analysis}.
    For ReCamMaster, regions that are incorrectly synthesized in the generated video coincide with attention that is scattered and mislocalized in {both} the reference and generated views, revealing a breakdown of cross-view correspondence.
    In contrast, our method produces substantially higher-fidelity generations while keeping the attention concentrated and well-aligned with the true corresponding region of the query point.}
    \label{supple_fig:attention_map_vis_in_davis_recam}
\end{figure*}

We examine the relationship between generation fidelity and the model's internal attention behavior
As established there, the 18th layer acts as the \emph{correspondence-specialized layer}, where each query point most strongly attends to its geometrically corresponding location across views; we therefore visualize the attention map at this layer.
We find that synthesis errors are tightly coupled with a breakdown of this alignment: when a query point is placed on an incorrectly generated region, a model without multi-view correspondence supervision attends to spurious locations in {both} the reference view and the generated view, rather than to the physically corresponding point.
As shown in Fig.~\ref{supple_fig:attention_map_vis_in_davis_recam}, ReCamMaster~\cite{bai2025recammaster} produces visible geometric inconsistency in the dynamic object, and its attention is dispersed onto irrelevant background structures instead of the queried object.
Built upon the same backbone, \ours\ not only synthesizes markedly higher-fidelity novel-view frames but also keeps the attention sharply concentrated and consistently aligned with the correct correspondences.
This visual evidence is consistent with our central finding that explicitly strengthening multi-view correspondence directly translates into geometrically faithful novel-view generation.

\section{Limitations}
While \ours\ achieves state-of-the-art geometric consistency and camera-pose accuracy on dynamic novel-view generation, several limitations remain. First, our method inherits the architectural constraints of the underlying ReCamMaster~\cite{bai2025recammaster} backbone, including a fixed output resolution of $480{\times}832$ and a maximum sequence length of 81 frames, which restricts its applicability to longer or higher-resolution videos. Second, the auxiliary tracking head requires ground-truth multi-view point correspondences during training, which currently restricts training to datasets where such supervision is available (e.g., synthetic or multi-camera captures); extending to fully unconstrained in-the-wild videos would require pseudo-labeling or self-supervised correspondence learning. Finally, our method shares the common limitation of diffusion-based video generation in terms of inference cost, which currently prevents real-time applications. We leave addressing these limitations to future work.

\clearpage
\clearpage


\clearpage
\clearpage

\clearpage
\clearpage

\clearpage
\clearpage

\begin{figure*}[t]
    \centering
    \includegraphics[width=\linewidth]{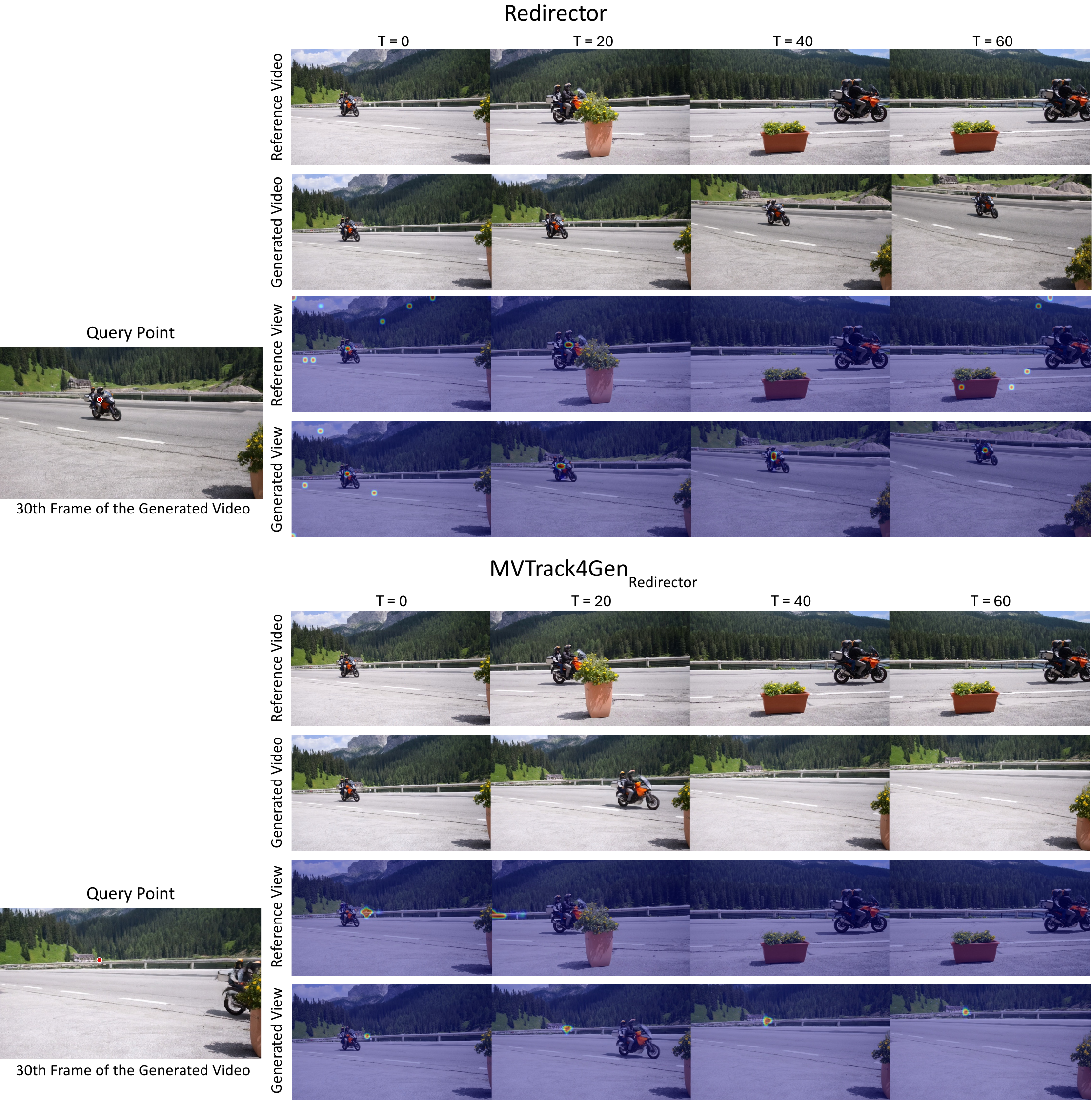}
    \caption{\textbf{Correlation between generation quality and attention-map alignment.}
    For both ReCamMaster~\cite{bai2025recammaster} (top) and our \ours\ (bottom), we show, from top to bottom, the {reference video}, the generated {novel-view video}, and the corresponding attention maps overlaid on the \emph{reference view} and the \emph{generated view} across frames.
    Given a query point placed on the 30th frame of the generated video (left), we visualize the attention distribution extracted from the 18th layer, identified as the correspondence-specialized layer in Sec.~\ref{sec:analysis}.
    For Redirector, regions that are incorrectly synthesized in the generated video coincide with attention that is scattered and mislocalized in {both} the reference and generated views, revealing a breakdown of cross-view correspondence.
    In contrast, our method produces substantially higher-fidelity generations while keeping the attention concentrated and well-aligned with the true corresponding region of the query point.}
    \label{supple_fig:attention_map_vis_in_davis_redirector}
\end{figure*}





\end{document}